\pdfoutput=1

\documentclass[11pt]{article}

\usepackage[final]{EMNLP2023}

\usepackage{times}
\usepackage{latexsym}

\usepackage[T1]{fontenc}

\usepackage[utf8]{inputenc}

\usepackage{microtype}

\usepackage{inconsolata}

\usepackage{enumitem}

\usepackage{CJKutf8}

\usepackage{listings}
\usepackage{color}

\usepackage{lscape}

\usepackage{import}

\usepackage{afterpage}

\usepackage{makecell}

\newcommand{\chinese}[1]{\begin{CJK}{UTF8}{gbsn}#1\end{CJK}}

\definecolor{dkgreen}{rgb}{0,0.6,0}
\definecolor{gray}{rgb}{0.5,0.5,0.5}
\definecolor{mauve}{rgb}{0.58,0,0.82}

\usepackage[counterclockwise, figuresleft]{rotating}

\newcommand{\lrot}[1]{\begin{turn}{270}#1\enspace\end{turn}}

\title{Mitigating Metric Bias in Minimum Bayes Risk Decoding}

\author{Geza Kovacs \and Daniel Deutsch \and Markus Freitag \\
  Google \\
  \texttt{\{geza, dandeutsch, freitag\}@google.com}
}

\begin{document}
\maketitle


\begin{abstract}

While Minimum Bayes Risk (MBR) decoding using metrics such as COMET or MetricX has outperformed traditional decoding methods such as greedy or beam search, it introduces a challenge we refer to as \emph{metric bias}. As MBR decoding aims to produce translations that score highly according to a specific utility metric, this very process makes it impossible to use the same metric for both decoding and evaluation, as improvements might simply be due to \emph{reward hacking} rather than reflecting real quality improvements. In this work we find that compared to human ratings, neural metrics not only overestimate the quality of MBR decoding when the same metric is used as the utility metric, but they also overestimate the quality of MBR/QE decoding with other neural utility metrics as well. We also show that the metric bias issue can be mitigated by using an ensemble of utility metrics during MBR decoding: human evaluations show that MBR decoding using an ensemble of utility metrics outperforms a single utility metric.

\end{abstract}

\section{Introduction}

Minimum bayes risk (MBR) decoding is a decoding approach where $n$ candidate translations are sampled from the MT system, and they are used as pseudoreferences for a reference-based \textit{utility metric}. MBR decoding computes the utility metric for all $O(n^2)$ pairs of candidates and pseudoreferences, selecting the candidate that achieves the best average score across all pseudoreferences. Quality Estimation (QE) decoding\footnote{Also known as QE reranking or QE filtering.} selects the candidate that scores best according to a QE utility metric. Previous work on MBR decoding has shown that it results in improvements on the utility metric~\citep{amrhein-sennrich-2022-identifying, cheng-vlachos-2023-faster, eikema-aziz-2022-sampling}, however other metrics do not improve as much as the utility metric~\citep{guttmann2024chasing, vamvas2024linear}. This issue of MBR/QE decoding exhibiting bias towards the utility metric complicates our ability to use automatic metrics to compare the quality of MBR/QE-based MT systems, as we cannot tell whether improvements in automatic metrics from MBR/QE decoding correspond to actual improvements in quality, or if it simply reward hacking. Prior work has assumed that this issue can be avoided by using a different metric for evaluating MBR decoding outputs~\citep{tomani2023quality}, though this assumption has never been tested.

In this work we compare the results of human vs metric-based evaluation of MBR/QE decoding with a wide variety of metrics to show that the quality of MBR/QE decoding is overestimated by not only the utility metric, but also other similar metrics.
While MBR/QE decoding with a single utility metric results in significant gains in automatic metrics, it does not perform better than greedy decoding in our human evaluations. 
This may be due to MBR decoding preferring fluent yet inaccurate candidates. Using an ensemble of metrics as the utility helps us mitigate the metric bias issue, with human evaluations showing that MBR decoding with an ensemble utility metric results in significantly better translations than greedy decoding or MBR/QE decoding with a single utility metric.

In this paper we contribute:

\begin{enumerate}[topsep=0pt,itemsep=-1ex,partopsep=1ex,parsep=1ex]
\item A large-scale analysis of metric bias in MBR and QE decoding with metrics commonly used in MT, showing that this metric bias issue holds across many different metrics and language pairs, and is not resolved by simply using a different metric for evaluation.
\item Mitigation strategies for MBR bias using QE filtering followed by MBR decoding, as well as MBR decoding using an ensemble of metrics as the utility function.
\item A human evaluation showing that MBR decoding with ensembles outperforms MBR decoding with a single metric.
\end{enumerate}


\section{Related Work}


\citet{cheng-vlachos-2023-faster, eikema-aziz-2022-sampling, guttmann2024chasing} find that MBR decoding improves automated metrics on various high, medium, and low resource language pairs. \citet{freitag-etal-2023-epsilon, freitag-etal-2022-high, tomani2023quality} find that human raters prefer the outputs of MBR/QE decoding over greedy decoding.

MBR variants achieve speedups via heuristics ~\citep{trabelsi2024efficient, jinnai2024hyperparameter}, filtering pseudoreferences via a QE metric ~\citep{deguchi2024centroid, deguchi-etal-2023-naist} or filtering via another reference-based metric ~\citep{vamvas2024linear, eikema-aziz-2022-sampling}. 
Quality-aware translation, which incorporates quality estimation into the training process, has been found to improve translation quality over standard MBR~\citep{tomani2023quality}.

Other techniques for aligning translation models with human preferences include direct preference optimization~\citep{rafailov2024direct, yang-etal-2024-direct}, reinforcement learning from human feedback~\citep{christiano2017deep}, and reinforcement learning from AI feedback~\citep{bai2022constitutional}. 

~\citet{guttmann2024chasing, vamvas2024linear} show evidence of metric bias in MBR decoding, as they find that neural evaluation metrics favor models using MBR on the metric used as the utility function. However, these papers only cover only 2 metrics, and neither have human evaluations.

~\citet{sellam-etal-2020-learning, freitag-etal-2023-results, glushkova-etal-2023-bleu} find that ensembling metrics can improve their ability to detect critical errors and improve agreement with human preferences, though they do not investigate the effects of ensembling utility metrics on MBR decoding.

Reward hacking ~\citep{skalse2022defining} is an issue in reinforcement learning where the reward function improves but the system's behavior is not aligned with human preferences. The metric bias problem in MBR decoding can be viewed as an instance of reward hacking, as the utility function improves while not necessarily improving quality.

\section{Study 1: Metric Bias in MBR Decoding}


\subsection{Methodology}

To investigate metric bias in MBR/QE decoding, we perform MBR/QE decoding via various utility metrics and compare how they perform on various evaluation metrics. We investigate MBR decoding using these reference-based utility metrics:

\begin{enumerate}[topsep=0pt,itemsep=-1ex,partopsep=1ex,parsep=1ex]
\item MetricX-23~\citep{juraska-etal-2023-metricx}
\item XCOMET-XXL~\citep{guerreiro2023xcomet}
\item XCOMET-XL~\citep{guerreiro2023xcomet}
\item COMET22~\citep{rei-etal-2022-comet}
\item AfriCOMET~\citep{wang-etal-2024-afrimte}
\item IndicCOMET~\citep{sai-b-etal-2023-indicmt}
\item BLEURT~\citep{sellam-etal-2020-bleurt}
\item YiSi-1~\citep{lo-2019-yisi}
\item sentBLEU~\citep{papineni-etal-2002-bleu}
\item chrF~\citep{popovic-2015-chrf}
\item chrF++~\citep{popovic-2017-chrf}
\item TER~\citep{snover-etal-2006-study}
\end{enumerate}

We also investigate QE decoding~\citep{fernandes-etal-2022-quality} using the following QE metrics:

\begin{enumerate}[topsep=0pt,itemsep=-1ex,partopsep=1ex,parsep=1ex]
\item MetricX-QE~\citep{juraska-etal-2023-metricx}
\item CometKiwi23-XXL~\citep{rei-etal-2023-scaling}
\item CometKiwi23-XL~\citep{rei-etal-2023-scaling}
\item CometKiwi22~\citep{rei-etal-2022-cometkiwi}
\item AfriCOMET-QE~\citep{wang-etal-2024-afrimte}
\end{enumerate}


We used a dev set for selecting ensembles, and a test set for reporting final results and human evaluation. The dev datasets and language pairs are:



\begin{enumerate}[topsep=0pt,itemsep=-1ex,partopsep=1ex,parsep=1ex]

\item FLORES-200 dev set~\citep{costa2022no}: English-Swahili (en-sw), Igbo (en-ig), Hindi (en-hi), Tamil (en-ta), Somali (en-so), Hausa (en-ha), Malayalam (en-ml), Gujarati (en-gu), Hungarian (en-hu), Vietnamese (en-vi)
\item WMT2022~\citep{kocmi-etal-2022-findings}: English-Chinese (en-zh), Chinese-English (zh-en), English-German (en-de), German-English (de-en)
\end{enumerate}

The test set datasets and language pairs are:

\begin{enumerate}[topsep=0pt,itemsep=-1ex,partopsep=1ex,parsep=1ex]
\item FLORES-200 test set: en-sw, en-ig, en-hi, en-ta, en-so, en-ha, en-ml, en-gu, en-hu, en-vi
\item WMT2023~\citep{kocmi-etal-2023-findings}: en-zh, zh-en\footnote{Due to errors in the WMT2023 zh-en reference translations, we use the references from~\citet{liu2024beyond} for zh-en.}, en-de, de-en
\end{enumerate}

We produced translations using Gemini 1.0 Pro~\citep{team2023gemini} with prompts including 5-shot examples. We used epsilon sampling as recommended by~\citet{freitag-etal-2023-epsilon} with a sample size of 128. See Appendix~\ref{sec:methodologydetails} for prompts used for generating translations and instructions for computing scores from metrics.

\clearpage

\begin{table*}

\vspace{-15mm}

\makebox[\textwidth]{
\setlength\tabcolsep{2 pt}
\begin{small}

\begin{tabular}{lcccccccccccccc}
\textbf{MBR/QE Method} \hfill \hfill \hfill \hfill \hfill \hfill \hfill \hfill \hfill \hfill \lrot{\textbf{\scriptsize Evaluated Metric}} & \lrot{MetricX} & \lrot{MetricX-QE} & \lrot{XCOMET-XXL} & \lrot{XCOMET-XL} & \lrot{\scriptsize CometKiwi23-XXL} & \lrot{\scriptsize CometKiwi23-XL} & \lrot{CometKiwi22} & \lrot{COMET22} & \lrot{BLEURT} & \lrot{YiSi} & \lrot{chrF} & \lrot{chrF++} & \lrot{sentBLEU} & \lrot{TER} \\
Greedy & {\cellcolor[HTML]{FDFFFD}} \color[HTML]{000000} 1.58 & {\cellcolor[HTML]{FDFFFD}} \color[HTML]{000000} 1.16 & {\cellcolor[HTML]{FDFFFD}} \color[HTML]{000000} 82.3 & {\cellcolor[HTML]{FDFFFD}} \color[HTML]{000000} 77.8 & {\cellcolor[HTML]{FDFFFD}} \color[HTML]{000000} 76.8 & {\cellcolor[HTML]{FDFFFD}} \color[HTML]{000000} 68.2 & {\cellcolor[HTML]{FDFFFD}} \color[HTML]{000000} 77.5 & {\cellcolor[HTML]{FDFFFD}} \color[HTML]{000000} 85.2 & {\cellcolor[HTML]{FDFFFD}} \color[HTML]{000000} 77.3 & {\cellcolor[HTML]{FDFFFD}} \color[HTML]{000000} 84.3 & {\cellcolor[HTML]{FDFFFD}} \color[HTML]{000000} 57.2 & {\cellcolor[HTML]{FDFFFD}} \color[HTML]{000000} 54.2 & {\cellcolor[HTML]{FDFFFD}} \color[HTML]{000000} 26.4 & {\cellcolor[HTML]{FDFFFD}} \color[HTML]{000000} 63.4 \\
\end{tabular}

\end{small}
}
\begin{small}

\caption{Reference-based and QE evaluation scores for greedy and MBR/QE decoding (1\textsuperscript{st} block), and ensembles (2\textsuperscript{nd} block), averaged across all languages (test datasets). Higher scores are better, except MetricX, MetricX-QE, and TER, where lower is better. Green is better than greedy, red is worse.  Ensembles are defined in Table~\ref{tab:ensembles}. Significant differences from greedy (pairwise t-test) indicated by * for p<0.05, † for p<0.01, ‡ for p<0.001. The green diagonal in the 1\textsuperscript{st} block shows metrics prefer outputs from MBR/QE decoding using the same utility metric.}
\label{tab:subset_all_test}
\end{small}

\vspace{5mm}


\setlength\tabcolsep{5 pt}
\begin{small}
\begin{tabular}{|l|c|c|c|c|c|c|c|c|c|c|c|c|c|c|c|c|c|}
\hline
          & \lrot{MetricX-QE}                                      & \lrot{\scriptsize CometKiwi23-XXL}          & \lrot{\scriptsize CometKiwi23-XL}           & \lrot{CometKiwi22}              & \lrot{\vtop{\hbox{\strut \scriptsize AfriCOMET-QE}\hbox{\strut (African only)}}} & \lrot{MetricX}                  & \lrot{XCOMET-XXL}                                      & \lrot{XCOMET-XL}                & \lrot{COMET22}                  & \lrot{\vtop{\hbox{\strut AfriCOMET}\hbox{\strut (African only)}}} & \lrot{\vtop{\hbox{\strut IndicCOMET}\hbox{\strut (Indic only)}}} & \lrot{BLEURT}                   & \lrot{YiSi}                     & \lrot{chrF}                 & \lrot{chrF++}                     & \lrot{sentBLEU}                   & \lrot{TER}                      \\ \hline
all       & \cellcolor[HTML]{000000}{\color[HTML]{000000} } & \cellcolor[HTML]{000000} & \cellcolor[HTML]{000000} & \cellcolor[HTML]{000000} & \cellcolor[HTML]{000000}             & \cellcolor[HTML]{000000} & \cellcolor[HTML]{000000}{\color[HTML]{000000} } & \cellcolor[HTML]{000000} & \cellcolor[HTML]{000000} & \cellcolor[HTML]{000000}          & \cellcolor[HTML]{000000}        & \cellcolor[HTML]{000000} & \cellcolor[HTML]{000000} & \cellcolor[HTML]{000000} & \cellcolor[HTML]{000000} & \cellcolor[HTML]{000000} & \cellcolor[HTML]{000000} \\ \hline
qe        & \cellcolor[HTML]{000000}                        & \cellcolor[HTML]{000000} & \cellcolor[HTML]{000000} & \cellcolor[HTML]{000000} & \cellcolor[HTML]{000000}             &                          &                                                 &                          &                          &                                   &                                 &                          &                          &                          &                          &                          &                          \\ \hline
top       & \cellcolor[HTML]{000000}                        & \cellcolor[HTML]{000000} & \cellcolor[HTML]{000000} &                          &                                      & \cellcolor[HTML]{000000} & \cellcolor[HTML]{000000}                        & \cellcolor[HTML]{000000} &                          &                                   &                                 &                          &                          &                          &                          &                          &                          \\ \hline
topQe     & \cellcolor[HTML]{000000}                        & \cellcolor[HTML]{000000} & \cellcolor[HTML]{000000} &                          &                                      &                          &                                                 &                          &                          &                                   &                                 &                          &                          &                          &                          &                          &                          \\ \hline
mxmxqe    & \cellcolor[HTML]{000000}                        &                          &                          &                          &                                      & \cellcolor[HTML]{000000} &                                                 &                          &                          &                                   &                                 &                          &                          &                          &                          &                          &                          \\ \hline
noLex     & \cellcolor[HTML]{000000}                        & \cellcolor[HTML]{000000} & \cellcolor[HTML]{000000} & \cellcolor[HTML]{000000} & \cellcolor[HTML]{000000}             & \cellcolor[HTML]{000000} & \cellcolor[HTML]{000000}                        & \cellcolor[HTML]{000000} & \cellcolor[HTML]{000000} & \cellcolor[HTML]{000000}          & \cellcolor[HTML]{000000}        & \cellcolor[HTML]{000000} & \cellcolor[HTML]{000000} &                          &                          &                          &                          \\ \hline
noNC      & \cellcolor[HTML]{000000}                        &                          &                          &                          & \cellcolor[HTML]{000000}             & \cellcolor[HTML]{000000} &                                                 &                          & \cellcolor[HTML]{000000} & \cellcolor[HTML]{000000}          & \cellcolor[HTML]{000000}        & \cellcolor[HTML]{000000} & \cellcolor[HTML]{000000} & \cellcolor[HTML]{000000} & \cellcolor[HTML]{000000} & \cellcolor[HTML]{000000} & \cellcolor[HTML]{000000} \\ \hline
noNCnoLex & \cellcolor[HTML]{000000}                        &                          &                          &                          & \cellcolor[HTML]{000000}             & \cellcolor[HTML]{000000} &                                                 &                          & \cellcolor[HTML]{000000} & \cellcolor[HTML]{000000}          & \cellcolor[HTML]{000000}        & \cellcolor[HTML]{000000} & \cellcolor[HTML]{000000} &                          &                          &                          &                          \\ \hline
noNCQe    & \cellcolor[HTML]{000000}                        &                          &                          &                          & \cellcolor[HTML]{000000}             &                          &                                                 &                          &                          &                                   &                                 &                          &                          &                          &                          &                          &                          \\ \hline
allQE(N)allMBR        & \cellcolor[HTML]{009901}{\color[HTML]{FFFFFF} 1} & \cellcolor[HTML]{009901}{\color[HTML]{FFFFFF} 1} & \cellcolor[HTML]{009901}{\color[HTML]{FFFFFF} 1} & \cellcolor[HTML]{009901}{\color[HTML]{FFFFFF} 1} & \cellcolor[HTML]{009901}{\color[HTML]{FFFFFF} 1} & \cellcolor[HTML]{CB0000}{\color[HTML]{FFFFFF} 2} & \cellcolor[HTML]{CB0000}{\color[HTML]{FFFFFF} 2} & \cellcolor[HTML]{CB0000}{\color[HTML]{FFFFFF} 2} & \cellcolor[HTML]{CB0000}{\color[HTML]{FFFFFF} 2} & \cellcolor[HTML]{CB0000}{\color[HTML]{FFFFFF} 2} & \cellcolor[HTML]{CB0000}{\color[HTML]{FFFFFF} 2} & \cellcolor[HTML]{CB0000}{\color[HTML]{FFFFFF} 2} & \cellcolor[HTML]{CB0000}{\color[HTML]{FFFFFF} 2} & \cellcolor[HTML]{CB0000}{\color[HTML]{FFFFFF} 2} & \cellcolor[HTML]{CB0000}{\color[HTML]{FFFFFF} 2} & \cellcolor[HTML]{CB0000}{\color[HTML]{FFFFFF} 2} & \cellcolor[HTML]{CB0000}{\color[HTML]{FFFFFF} 2} \\ \hline
allQE(N)nolexMBR      & \cellcolor[HTML]{009901}{\color[HTML]{FFFFFF} 1} & \cellcolor[HTML]{009901}{\color[HTML]{FFFFFF} 1} & \cellcolor[HTML]{009901}{\color[HTML]{FFFFFF} 1} & \cellcolor[HTML]{009901}{\color[HTML]{FFFFFF} 1} & \cellcolor[HTML]{009901}{\color[HTML]{FFFFFF} 1} & \cellcolor[HTML]{CB0000}{\color[HTML]{FFFFFF} 2} & \cellcolor[HTML]{CB0000}{\color[HTML]{FFFFFF} 2} & \cellcolor[HTML]{CB0000}{\color[HTML]{FFFFFF} 2} & \cellcolor[HTML]{CB0000}{\color[HTML]{FFFFFF} 2} & \cellcolor[HTML]{CB0000}{\color[HTML]{FFFFFF} 2} & \cellcolor[HTML]{CB0000}{\color[HTML]{FFFFFF} 2} & \cellcolor[HTML]{CB0000}{\color[HTML]{FFFFFF} 2} & \cellcolor[HTML]{CB0000}{\color[HTML]{FFFFFF} 2} &                                                  &                                                  &                                                  &                                                  \\ \hline
topQE(N)topMBR        & \cellcolor[HTML]{009901}{\color[HTML]{FFFFFF} 1} & \cellcolor[HTML]{009901}{\color[HTML]{FFFFFF} 1} & \cellcolor[HTML]{009901}{\color[HTML]{FFFFFF} 1} & {\color[HTML]{FFFFFF} }                          & {\color[HTML]{FFFFFF} }                          & \cellcolor[HTML]{CB0000}{\color[HTML]{FFFFFF} 2} & \cellcolor[HTML]{CB0000}{\color[HTML]{FFFFFF} 2} & \cellcolor[HTML]{CB0000}{\color[HTML]{FFFFFF} 2} & {\color[HTML]{FFFFFF} }                          & {\color[HTML]{FFFFFF} }                          &                                                  &                                                  &                                                  &                                                  &                                                  &                                                  &                                                  \\ \hline
noncQE(N)noncMBR      & \cellcolor[HTML]{009901}{\color[HTML]{FFFFFF} 1} & \cellcolor[HTML]{FFFFFF}{\color[HTML]{FFFFFF} }  & \cellcolor[HTML]{FFFFFF}{\color[HTML]{FFFFFF} }  & \cellcolor[HTML]{FFFFFF}{\color[HTML]{FFFFFF} }  & \cellcolor[HTML]{009901}{\color[HTML]{FFFFFF} 1} & \cellcolor[HTML]{CB0000}{\color[HTML]{FFFFFF} 2} & \cellcolor[HTML]{FFFFFF}{\color[HTML]{FFFFFF} }  & \cellcolor[HTML]{FFFFFF}{\color[HTML]{FFFFFF} }  & \cellcolor[HTML]{CB0000}{\color[HTML]{FFFFFF} 2} & \cellcolor[HTML]{CB0000}{\color[HTML]{FFFFFF} 2} & \cellcolor[HTML]{CB0000}{\color[HTML]{FFFFFF} 2} & \cellcolor[HTML]{CB0000}{\color[HTML]{FFFFFF} 2} & \cellcolor[HTML]{CB0000}{\color[HTML]{FFFFFF} 2} & \cellcolor[HTML]{CB0000}{\color[HTML]{FFFFFF} 2} & \cellcolor[HTML]{CB0000}{\color[HTML]{FFFFFF} 2} & \cellcolor[HTML]{CB0000}{\color[HTML]{FFFFFF} 2} & \cellcolor[HTML]{CB0000}{\color[HTML]{FFFFFF} 2} \\ \hline
noncQE(N)noncnolexMBR & \cellcolor[HTML]{009901}{\color[HTML]{FFFFFF} 1} &                                                  &                                                  &                                                  & \cellcolor[HTML]{009901}{\color[HTML]{FFFFFF} 1} & \cellcolor[HTML]{CB0000}{\color[HTML]{FFFFFF} 2} & {\color[HTML]{FFFFFF} }                          & {\color[HTML]{FFFFFF} }                          & \cellcolor[HTML]{CB0000}{\color[HTML]{FFFFFF} 2} & \cellcolor[HTML]{CB0000}{\color[HTML]{FFFFFF} 2} & \cellcolor[HTML]{CB0000}{\color[HTML]{FFFFFF} 2} & \cellcolor[HTML]{CB0000}{\color[HTML]{FFFFFF} 2} & \cellcolor[HTML]{CB0000}{\color[HTML]{FFFFFF} 2} & {\color[HTML]{FFFFFF} }                          & {\color[HTML]{FFFFFF} }                          & {\color[HTML]{FFFFFF} }                          & {\color[HTML]{FFFFFF} }                          \\ \hline
mxQE(N)xcMBR          & \cellcolor[HTML]{009901}{\color[HTML]{FFFFFF} 1} &                                                  &                                                  &                                                  &                                                  &                                                  & \cellcolor[HTML]{CB0000}{\color[HTML]{FFFFFF} 2} &                                                  &                                                  &                                                  &                                                  &                                                  &                                                  &                                                  &                                                  &                                                  &                                                  \\ \hline
ckQE(N)xcMBR          &                                                  & \cellcolor[HTML]{009901}{\color[HTML]{FFFFFF} 1} &                                                  &                                                  &                                                  &                                                  & \cellcolor[HTML]{CB0000}{\color[HTML]{FFFFFF} 2} &                                                  &                                                  &                                                  &                                                  &                                                  &                                                  &                                                  &                                                  &                                                  &                                                  \\ \hline
mxQE(N)mxMBR          & \cellcolor[HTML]{009901}{\color[HTML]{FFFFFF} 1} &                                                  &                                                  &                                                  &                                                  & \cellcolor[HTML]{CB0000}{\color[HTML]{FFFFFF} 2} &                                                  &                                                  &                                                  &                                                  &                                                  &                                                  &                                                  &                                                  &                                                  &                                                  &                                                  \\ \hline
ckQE(N)mxMBR          &                                                  & \cellcolor[HTML]{009901}{\color[HTML]{FFFFFF} 1} &                                                  &                                                  &                                                  & \cellcolor[HTML]{CB0000}{\color[HTML]{FFFFFF} 2} &                                                  &                                                  &                                                  &                                                  &                                                  &                                                  &                                                  &                                                  &                                                  &                                                  &                                                  \\ \hline
\end{tabular}
\caption{Metrics included in each ensemble. Rows are ensembles, columns are metrics. Black cells indicate that the metric is included in a single-step ensemble. Green cells indicate the metric is used for the 1\textsuperscript{st} step (QE filtering) in a 2-step ensemble. Red cells indicate the metric is used for the 2\textsuperscript{nd} step (MBR decoding) in a 2-step ensemble.}
\label{tab:ensembles}
\end{small}

\end{table*}

\thispagestyle{empty}

\clearpage

\subsection{Results}

Results are shown in Table~\ref{tab:subset_all_test} for average scores across all language pairs on the test datasets. We observe that for all reference-based metrics, the best-performing system is MBR decoding using the same utility metric. This result also holds for all QE metrics, but that is by definition, because QE decoding picks the sample with the best QE score. These results also hold on individual languages and the dev set (Appendix~\ref{sec:perlangresults} and ~\ref{sec:devresults}).

We can also see that MBR decoding outputs for utility metrics which are similar to the evaluation metric tend to score better than when the MBR utility metric is dissimilar to the evaluation metric. For example, MBR/QE decoding with neural metrics (MetricX and COMET families) performs better than greedy when evaluated with other neural metrics, but worse than greedy if evaluated via lexical metrics. Likewise, MBR decoding with lexical metrics (sentBLEU, chrF, chrF++, and TER) and semantic metrics (YiSi) perform highly when evaluated by lexical and semantic metrics, but poorly when evaluated via neural metrics. The pattern also holds for similar metrics within the same family -- XCOMET-XXL prefers MBR/QE decoding using CometKiwi23-XXL and XCOMET-XL, and MetricX prefers outputs from MetricX-QE. 

These results suggest the existence of metric bias in MBR decoding -- that is, they suggest that MBR decoding will result in a disproportionately large improvement in the utility metric and metrics similar to the utility metric, relative to the actual improvement in quality. In order to address this issue, in the next section we will investigate ensembling metrics during MBR decoding as a means of avoiding overfitting to a particular utility metric.

\section{Study 2: MBR Decoding using Ensembles of Metrics}

\subsection{Methodology}

As a mitigation strategy for utility metric bias in MBR decoding, we investigate how using an ensemble of metrics performs for MBR decoding. We explore the following ensembling techniques (see Appendix~\ref{sec:ensemblecode} for pseudocode for these techniques):

\begin{enumerate}[topsep=0pt,itemsep=-1ex,partopsep=1ex,parsep=1ex]
\item rankAvg: For each metric, assigns a rank to each of the 128 samples (where 0 is best and 127 is worst). Select the sample where the average rank across metrics is minimized.
\item rankMed: Select the sample where the median rank across metrics is minimized.
\item rankMax: Select the sample where the maximum rank across metrics is minimized.
\item rank75q: Select the sample where the 0.75th quartile rank across metrics is minimized.
\end{enumerate}

For each of these ensembling techniques, we compute ensembles with the following groups of metrics (see Table~\ref{tab:ensembles} and Appendix~\ref{sec:ensemblemetrics} for the complete list of metrics included in each ensemble):

\begin{enumerate}[topsep=0pt,itemsep=-1ex,partopsep=1ex,parsep=1ex]
\item all: All metrics
\item qe: All QE metrics
\item top: Top-performing metrics in WMT2023 metrics shared task~\citep{freitag-etal-2023-results}
\item topQe: Top-performing QE metrics
\item mxmxqe: MetricX + MetricX-QE ensemble
\item noLex: Non-lexical metrics
\item noNC: Metrics that permit commercial use
\item noNCnoLex: Non-lexical metrics that permit commercial use
\item noNCQe: QE metrics that permit commercial use
\end{enumerate}

In addition to the ensembles above, we also investigate QE filtering followed by MBR decoding (QE filtering selects the top N candidates according to a QE metric, where N can be either 4, 8, 16, 32, 64). This two-step approach is faster than standard MBR decoding, as QE filtering is linear-time whereas MBR decoding is quadratic time. We include the following two-step ensembles:

\begin{enumerate}[topsep=0pt,itemsep=-1ex,partopsep=1ex,parsep=1ex]
\item allQE(N)allMBR: QE filter with all QE metrics, then MBR decode with all reference-based metrics
\item allQE(N)nolexMBR: QE filter with all QE metrics, then MBR decode with non-lexical reference-based metrics
\item topQE(N)topMBR: QE filter with top QE metrics, then MBR decode with top reference-based metrics
\item noncQE(N)noncMBR: QE filter with QE metrics that permit commercial use, then MBR decode with reference-based metrics that permit commercial use
\item noncQE(N)noncnolexMBR: QE filter with QE metrics that permit commerical use, then MBR decode with non-lexical reference-based metrics that permit commercial use
\item mxQE(N)xcMBR: QE filter with MetricX-QE, then MBR decode with XCOMET-XXL
\item ckQE(N)xcMBR: QE filter with CometKiwi23-XXL, then MBR decode with XCOMET-XXL
\item mxQE(N)mxMBR: QE filter with MetricX-QE, then MBR decode with MetricX
\item ckQE(N)mxMBR: QE filter with CometKiwi23-XXL, then MBR decode with MetricX
\end{enumerate}

The metrics included in each ensemble is shown in Table~\ref{tab:ensembles} and Appendix~\ref{sec:ensemblemetrics}.

\subsection{Results}

Results for a subset of ensembles averaged across all language pairs on the test sets are at Table~\ref{tab:subset_all_test} with additional ensembles shown in Appendix~\ref{sec:extraensembles}. Results on the dev sets are shown in Appendix~\ref{sec:devresults}. Breakdowns per language pair can be found in Appendix~\ref{sec:perlangresults}. As expected, ensembles tend to perform better if judged by metrics that are better represented in the ensemble; for example, if judging by MetricX, the best ensembles are mxQE(32)mxMBR and rankAvg:mxmxqe, both of which are ensembles consisting of MetricX and MetricX-QE.

That said, observe that compared to MBR/QE decoding with a single utility metric, ensembles often improve on automated evaluations even according to metrics not included in the ensemble. For example, if we use the XCOMET or CometKiwi families of metrics to evaluate rankAvg:noNCnoLex and noncQE(32)noncnolexMBR (which do not include any metrics from the XCOMET or CometKiwi families), they outperform MBR/QE decoding with any single metric outside the XCOMET or CometKiwi families. Similarly, if lexical metrics are used to evaluate the rankAvg:noLex and allQE(32)nolexMBR ensembles, which do not include any lexical metrics, they still outperform MBR/QE decoding with any single neural metric. This suggests that ensembles help reduce metric bias towards a single metric, which results in improved automated evaluation scores according to other metrics not included in the ensemble. 


\begin{table*}[!htb]
\begin{small}
\setlength\tabcolsep{2pt}

\begin{tabular}{lccccccccccccccc}
 & \lrot{Greedy} & \lrot{Reference} & \lrot{MetricX} & \lrot{MetricX-QE} & \lrot{XCOMET-XXL} & \lrot{CometKiwi23-XXL} & \lrot{COMET22} & \lrot{AfriCOMET} & \lrot{AfriCOMET-QE} & \lrot{IndicCOMET} & \lrot{rankAvg:all} & \lrot{rankAvg:noNC} & \lrot{\scriptsize rankAvg:noNCnoLex} & \lrot{mxQE32mxMBR} & \lrot{\scriptsize noncQE32noncnolexMBR} \\
 &  &  &  &  &  &  &  &  &  &  &  &  &  &  &  \\
all:total & {\cellcolor[HTML]{FFFDFD}} \color[HTML]{000000} 1.52 & {\cellcolor[HTML]{8B0000}} \color[HTML]{F1F1F1} \bfseries 1.80† & {\cellcolor[HTML]{F17979}} \color[HTML]{F1F1F1} 1.59 & {\cellcolor[HTML]{CF0000}} \color[HTML]{F1F1F1} \bfseries 1.77† & {\cellcolor[HTML]{000000}} \color[HTML]{F1F1F1}  & {\cellcolor[HTML]{000000}} \color[HTML]{F1F1F1}  & {\cellcolor[HTML]{000000}} \color[HTML]{F1F1F1}  & {\cellcolor[HTML]{000000}} \color[HTML]{F1F1F1}  & {\cellcolor[HTML]{000000}} \color[HTML]{F1F1F1}  & {\cellcolor[HTML]{000000}} \color[HTML]{F1F1F1}  & {\cellcolor[HTML]{000000}} \color[HTML]{F1F1F1}  & {\cellcolor[HTML]{006500}} \color[HTML]{F1F1F1} \bfseries 1.27‡ & {\cellcolor[HTML]{006800}} \color[HTML]{F1F1F1} \bfseries 1.28‡ & {\cellcolor[HTML]{FEF8F8}} \color[HTML]{000000} 1.53 & {\cellcolor[HTML]{006400}} \color[HTML]{F1F1F1} \bfseries 1.27‡ \\
\end{tabular}

\end{small}
\caption{Human evaluation results broken down by language and MQM error type. Columns indicate the system used for MBR/QE decoding; ensembles are defined in Table~\ref{tab:ensembles}. Rows starting with ``all'' shows results across all languages. 1\textsuperscript{st} block is total error scores, 2\textsuperscript{nd} is fluency error scores, 3\textsuperscript{rd} is accuracy error scores, 4\textsuperscript{th} is other error scores. For each system, average human evaluation scores across the evaluated segments are shown. Lower scores are better. Colors are relative to greedy, green is better than greedy, red is worse. Black cells were not evaluated. Significant differences from greedy (pairwise t-test) indicated by * for p<0.05, † for p<0.01, ‡ for p<0.001.}
\label{tab:human_eval_error_types}
\end{table*}

\begin{table*}[!htb]
\begin{small}
\setlength\tabcolsep{2pt}
\begin{tabular}{llcccccccc}
 System & Translation & \lrot{Fluency MQM} & \lrot{Accuracy MQM} & \lrot{Other MQM} & \lrot{MetricX} & \lrot{MetricX-QE} & \lrot{XCOMET-XXL} & \lrot{\scriptsize{CometKiwi23-XXL}} & \lrot{COMET22} \\
\hline
Greedy & \vtop{\hbox{\strut The seller said not yet, and it}\hbox{\strut \textbf{will} be shipped in the afternoon.}} & {\cellcolor[HTML]{FDFFFD}} \color[HTML]{000000} 1.0 & {\cellcolor[HTML]{FDFFFD}} \color[HTML]{000000} 0.0 & {\cellcolor[HTML]{FDFFFD}} \color[HTML]{000000} 0.0 & {\cellcolor[HTML]{FDFFFD}} \color[HTML]{000000} 0.659 & {\cellcolor[HTML]{FDFFFD}} \color[HTML]{000000} 0.88 & {\cellcolor[HTML]{FDFFFD}} \color[HTML]{000000} 0.999 & {\cellcolor[HTML]{FDFFFD}} \color[HTML]{000000} 0.83 & {\cellcolor[HTML]{FDFFFD}} \color[HTML]{000000} 0.74 \\
\hline
\vtop{\hbox{\strut MetricX}\hbox{\strut /XCOMET-XXL}} & \vtop{\hbox{\strut The seller said that they \textbf{don't}}\hbox{\strut \textbf{have it in stock yet}, and \textbf{will} be}\hbox{\strut able to ship it out \textbf{this afternoon}.}} & {\cellcolor[HTML]{FDFFFD}} \color[HTML]{000000} 1.0 & {\cellcolor[HTML]{8B0000}} \color[HTML]{F1F1F1} 10.0 & {\cellcolor[HTML]{FDFFFD}} \color[HTML]{000000} 0.0 & {\cellcolor[HTML]{006400}} \color[HTML]{F1F1F1} 0.259 & {\cellcolor[HTML]{F46060}} \color[HTML]{F1F1F1} 0.94 & {\cellcolor[HTML]{006400}} \color[HTML]{F1F1F1} 1.000 & {\cellcolor[HTML]{8B0000}} \color[HTML]{F1F1F1} 0.70 & {\cellcolor[HTML]{8B0000}} \color[HTML]{F1F1F1} 0.68 \\
\hline
MetricX-QE & \vtop{\hbox{\strut The seller said he hadn't shipped it,}\hbox{\strut but could ship it that afternoon.}} & {\cellcolor[HTML]{006400}} \color[HTML]{F1F1F1} 0.0 & {\cellcolor[HTML]{FDFFFD}} \color[HTML]{000000} 0.0 & {\cellcolor[HTML]{FDFFFD}} \color[HTML]{000000} 0.0 & {\cellcolor[HTML]{3FB33F}} \color[HTML]{F1F1F1} 0.438 & {\cellcolor[HTML]{006400}} \color[HTML]{F1F1F1} 0.49 & {\cellcolor[HTML]{8B0000}} \color[HTML]{F1F1F1} 0.997 & {\cellcolor[HTML]{F45E5E}} \color[HTML]{F1F1F1} 0.78 & {\cellcolor[HTML]{9A0000}} \color[HTML]{F1F1F1} 0.68 \\
\hline
CometKiwi23-XXL & \vtop{\hbox{\strut The seller said that it was not ready yet}\hbox{\strut and that it would be shipped that afternoon.}} & {\cellcolor[HTML]{006400}} \color[HTML]{F1F1F1} 0.0 & {\cellcolor[HTML]{FDFFFD}} \color[HTML]{000000} 0.0 & {\cellcolor[HTML]{FDFFFD}} \color[HTML]{000000} 0.0 & {\cellcolor[HTML]{006500}} \color[HTML]{F1F1F1} 0.264 & {\cellcolor[HTML]{43B643}} \color[HTML]{F1F1F1} 0.67 & {\cellcolor[HTML]{F74848}} \color[HTML]{F1F1F1} 0.998 & {\cellcolor[HTML]{006400}} \color[HTML]{F1F1F1} 0.87 & {\cellcolor[HTML]{F49E9E}} \color[HTML]{000000} 0.73 \\
\hline
COMET22 & \vtop{\hbox{\strut The seller said not \textbf{yet, it}}\hbox{\strut will be sent in the afternoon.}} & {\cellcolor[HTML]{FDFFFD}} \color[HTML]{000000} 1.0 & {\cellcolor[HTML]{FDFFFD}} \color[HTML]{000000} 0.0 & {\cellcolor[HTML]{FDFFFD}} \color[HTML]{000000} 0.0 & {\cellcolor[HTML]{8B0000}} \color[HTML]{F1F1F1} 0.981 & {\cellcolor[HTML]{8B0000}} \color[HTML]{F1F1F1} 1.06 & {\cellcolor[HTML]{FB2020}} \color[HTML]{F1F1F1} 0.998 & {\cellcolor[HTML]{199419}} \color[HTML]{F1F1F1} 0.86 & {\cellcolor[HTML]{3BAF3B}} \color[HTML]{F1F1F1} 0.76 \\
\hline
\vtop{\hbox{\strut noncQE32noncnolexMBR}\hbox{\strut /rankAvg:noNCnoLex}} & \vtop{\hbox{\strut The seller \textbf{said no, it} won't be}\hbox{\strut shipped until this afternoon.}} & {\cellcolor[HTML]{FDFFFD}} \color[HTML]{000000} 1.0 & {\cellcolor[HTML]{FDFFFD}} \color[HTML]{000000} 0.0 & {\cellcolor[HTML]{8B0000}} \color[HTML]{F1F1F1} 1.0 & {\cellcolor[HTML]{86ED86}} \color[HTML]{000000} 0.552 & {\cellcolor[HTML]{118E11}} \color[HTML]{F1F1F1} 0.60 & {\cellcolor[HTML]{FA2E2E}} \color[HTML]{F1F1F1} 0.998 & {\cellcolor[HTML]{F83A3A}} \color[HTML]{F1F1F1} 0.76 & {\cellcolor[HTML]{006400}} \color[HTML]{F1F1F1} 0.77 \\
\hline
 \vtop{\hbox{\strut rankAvg:noNC}\hbox{\strut /rankAvg:all}} & \vtop{\hbox{\strut The seller said not \textbf{yet,}}\hbox{\strut \textbf{it} \textbf{will} be shipped in the afternoon.}} & {\cellcolor[HTML]{8B0000}} \color[HTML]{F1F1F1} 2.0 & {\cellcolor[HTML]{FDFFFD}} \color[HTML]{000000} 0.0 & {\cellcolor[HTML]{FDFFFD}} \color[HTML]{000000} 0.0 & {\cellcolor[HTML]{BCFCBC}} \color[HTML]{000000} 0.608 & {\cellcolor[HTML]{F8C6C6}} \color[HTML]{000000} 0.90 & {\cellcolor[HTML]{FC1E1E}} \color[HTML]{F1F1F1} 0.998 & {\cellcolor[HTML]{56C656}} \color[HTML]{000000} 0.84 & {\cellcolor[HTML]{F84040}} \color[HTML]{F1F1F1} 0.71 \\
\hline
mxQE32mxMBR & \vtop{\hbox{\strut The seller said that it is not yet ready,}\hbox{\strut and it will be shipped in the afternoon.}} & {\cellcolor[HTML]{006400}} \color[HTML]{F1F1F1} 0.0 & {\cellcolor[HTML]{F93434}} \color[HTML]{F1F1F1} 5.0 & {\cellcolor[HTML]{FDFFFD}} \color[HTML]{000000} 0.0 & {\cellcolor[HTML]{3BAF3B}} \color[HTML]{F1F1F1} 0.432 & {\cellcolor[HTML]{76E076}} \color[HTML]{000000} 0.75 & {\cellcolor[HTML]{F93232}} \color[HTML]{F1F1F1} 0.998 & {\cellcolor[HTML]{6ED96E}} \color[HTML]{000000} 0.84 & {\cellcolor[HTML]{F39999}} \color[HTML]{000000} 0.73 \\
\end{tabular}
\end{small}
\caption{An example where MetricX and XCOMET-XXL MBR decoding result in an inaccurate translation. The source text is \chinese{卖家说还没，下午才能发。} (``Seller says not yet, can ship in the afternoon.'') The preceding sentence is \chinese{结果，第二天打电话问，发货了吗？} (``So the next day I called to ask, has it shipped?''). MetricX and XCOMET-XXL MBR decoding, as well as the reference-based MetricX and XCOMET-XXL evaluations, all prefer a translation which inaccurately states the item is out of stock. The other metrics assign a lower score to the inaccurate translation. Lower scores are better for MQM, MetricX, and MetricX-QE, for other metrics higher is better. Green is better than greedy, red is worse. Spans marked as errors by the rater are bolded.}
\label{tab:example_translations}
\end{table*}

\section{Study 3: Human Evaluation}

\subsection{Methodology}

For the human evaluation, we chose the following baselines and ensembles to evaluate:

\begin{enumerate}[topsep=0pt,itemsep=-1ex,partopsep=1ex,parsep=1ex]
\item Greedy decoding
\item Reference translation
\item MetricX (MBR decoding)
\item MetricX-QE (QE decoding)
\item AfriCOMET for African languages (MBR decoding)
\item AfriCOMET-QE for African languages (QE decoding)
\item IndicCOMET for Indic langauges (MBR decoding)
\item rankAvg:noNC (single-step ensemble)
\item rankAvg:noNCnoLex (single-step ensemble)
\item mxQE(32)mxMBR (multi-step ensemble)
\item noncQE(32)noncnolexMBR (multi-step ensemble)
\end{enumerate}

We evaluated the following conditions only on en-de and zh-en due to budget constraints:

\begin{enumerate}[topsep=0pt,itemsep=-1ex,partopsep=1ex,parsep=1ex]
\item XCOMET-XXL (MBR decoding)
\item CometKiwi23-XXL (QE decoding)
\item COMET22 (MBR decoding)
\item rankAvg:all (single-step ensemble)
\end{enumerate}

We chose MetricX, MetricX-QE, AfriCOMET, AfriCOMET-QE, and IndicCOMET because they had shown good performance in previously-published evaluations ~\citep{tomani2023quality, wang-etal-2024-afrimte, sai-b-etal-2023-indicmt, freitag-etal-2023-results}, had good performance in automated evaluations on the dev set (Appendix~\ref{sec:devresults}), and lacked restrictions on commercial use. In our en-de and zh-en evaluations we also included metrics and ensembles with restrictions on commercial use (XCOMET, CometKiwi, rankAvg:all) for comparison. The 6 language pairs and datasets we evaluate are en-ha en-sw en-ml en-hi (from FLORES200 test) and en-de zh-en (from WMT2023). We chose these languages to have a wide distribution in resource level. For each language pair, we sampled 400 source segments to evaluate. WMT2023 was evaluated with document context, whereas FLORES200 segments were evaluated in isolation. 
We asked each rater to provide MQM annotations for all translation candidates for each source segment (we evaluted 15 systems on en-de and zh-en and 11 systems on others), and compute scores as described in ~\citet{freitag-etal-2021-experts}. Scores range from 0 to 25, lower is better. To control for variance between raters, the same rater was used to score all candidate translations resulting from each source segment.




\begin{table*}[!htb]

\begin{small}
\setlength\tabcolsep{2pt}

\begin{tabular}{lllllllllllll}
 & \lrot{Greedy} & \lrot{Reference} & \lrot{MetricX} & \lrot{MetricX-QE} & \lrot{XCOMET-XXL} & \lrot{CometKiwi23-XXL} & \lrot{COMET22} & \lrot{rankAvg:all} & \lrot{rankAvg:noNC} & \lrot{\scriptsize rankAvg:noNCnoLex} & \lrot{mxQE32mxMBR} & \lrot{\scriptsize noncQE32noncnolexMBR} \\
 &  &  &  &  &  &  &  &  &  &  &  &  \\
en-de@news:total & {\cellcolor[HTML]{FFFDFD}} \color[HTML]{000000} 1.95 & {\cellcolor[HTML]{F93030}} \color[HTML]{F1F1F1} 2.97 & {\cellcolor[HTML]{E10000}} \color[HTML]{F1F1F1} 3.47 & {\cellcolor[HTML]{FE0B0B}} \color[HTML]{F1F1F1} 3.28 & {\cellcolor[HTML]{F36767}} \color[HTML]{F1F1F1} 2.52 & {\cellcolor[HTML]{8B0000}} \color[HTML]{F1F1F1} 3.74 & {\cellcolor[HTML]{FEF3F3}} \color[HTML]{000000} 1.99 & {\cellcolor[HTML]{FEF3F3}} \color[HTML]{000000} 1.99 & {\cellcolor[HTML]{F6B2B2}} \color[HTML]{000000} 2.16 & {\cellcolor[HTML]{80E880}} \color[HTML]{000000} 1.91 & {\cellcolor[HTML]{FBDFDF}} \color[HTML]{000000} 2.05 & {\cellcolor[HTML]{006400}} \color[HTML]{F1F1F1} 1.78 \\
en-de@user-review:total & {\cellcolor[HTML]{FFFDFD}} \color[HTML]{000000} 3.66 & {\cellcolor[HTML]{006E00}} \color[HTML]{F1F1F1} 2.79 & {\cellcolor[HTML]{64D164}} \color[HTML]{000000} 3.30 & {\cellcolor[HTML]{006F00}} \color[HTML]{F1F1F1} 2.80 & {\cellcolor[HTML]{2AA22A}} \color[HTML]{F1F1F1} 3.11 & {\cellcolor[HTML]{8B0000}} \color[HTML]{F1F1F1} 4.07 & {\cellcolor[HTML]{FA2C2C}} \color[HTML]{F1F1F1} 3.90 & {\cellcolor[HTML]{F5ADAD}} \color[HTML]{000000} 3.71 & {\cellcolor[HTML]{007000}} \color[HTML]{F1F1F1} 2.81 & {\cellcolor[HTML]{2DA42D}} \color[HTML]{F1F1F1} 3.12 & {\cellcolor[HTML]{239C23}} \color[HTML]{F1F1F1} 3.09 & {\cellcolor[HTML]{006400}} \color[HTML]{F1F1F1} 2.68 \\
en-de@mastodon:total & {\cellcolor[HTML]{FFFDFD}} \color[HTML]{000000} 1.29 & {\cellcolor[HTML]{FD1212}} \color[HTML]{F1F1F1} 1.70 & {\cellcolor[HTML]{5CCB5C}} \color[HTML]{000000} 1.17 & {\cellcolor[HTML]{F83A3A}} \color[HTML]{F1F1F1} 1.60 & {\cellcolor[HTML]{F83A3A}} \color[HTML]{F1F1F1} 1.60 & {\cellcolor[HTML]{8B0000}} \color[HTML]{F1F1F1} \bfseries 1.87* & {\cellcolor[HTML]{3BB03B}} \color[HTML]{F1F1F1} 1.13 & {\cellcolor[HTML]{74DE74}} \color[HTML]{000000} 1.19 & {\cellcolor[HTML]{007600}} \color[HTML]{F1F1F1} 1.04 & {\cellcolor[HTML]{007200}} \color[HTML]{F1F1F1} 1.03 & {\cellcolor[HTML]{FB2626}} \color[HTML]{F1F1F1} 1.65 & {\cellcolor[HTML]{006400}} \color[HTML]{F1F1F1} 0.98 \\
en-de@speech:total & {\cellcolor[HTML]{FFFDFD}} \color[HTML]{000000} 3.59 & {\cellcolor[HTML]{F7BCBC}} \color[HTML]{000000} 3.78 & {\cellcolor[HTML]{98FB98}} \color[HTML]{000000} 3.37 & {\cellcolor[HTML]{006E00}} \color[HTML]{F1F1F1} 2.60 & {\cellcolor[HTML]{8B0000}} \color[HTML]{F1F1F1} \bfseries 5.43* & {\cellcolor[HTML]{F5ADAD}} \color[HTML]{000000} 3.83 & {\cellcolor[HTML]{2AA22A}} \color[HTML]{F1F1F1} 2.97 & {\cellcolor[HTML]{2AA22A}} \color[HTML]{F1F1F1} 2.97 & {\cellcolor[HTML]{128F12}} \color[HTML]{F1F1F1} 2.88 & {\cellcolor[HTML]{007200}} \color[HTML]{F1F1F1} 2.65 & {\cellcolor[HTML]{006F00}} \color[HTML]{F1F1F1} 2.61 & {\cellcolor[HTML]{006400}} \color[HTML]{F1F1F1} 2.48 \\
zh-en@news:total & {\cellcolor[HTML]{FFFDFD}} \color[HTML]{000000} 3.56 & {\cellcolor[HTML]{8B0000}} \color[HTML]{F1F1F1} 4.51 & {\cellcolor[HTML]{F45D5D}} \color[HTML]{F1F1F1} 3.90 & {\cellcolor[HTML]{F45D5D}} \color[HTML]{F1F1F1} 3.90 & {\cellcolor[HTML]{F26E6E}} \color[HTML]{F1F1F1} 3.83 & {\cellcolor[HTML]{FB2424}} \color[HTML]{F1F1F1} 4.16 & {\cellcolor[HTML]{6BD66B}} \color[HTML]{000000} 3.36 & {\cellcolor[HTML]{006400}} \color[HTML]{F1F1F1} 2.98 & {\cellcolor[HTML]{F45F5F}} \color[HTML]{F1F1F1} 3.90 & {\cellcolor[HTML]{007F00}} \color[HTML]{F1F1F1} 3.15 & {\cellcolor[HTML]{F26F6F}} \color[HTML]{F1F1F1} 3.83 & {\cellcolor[HTML]{FA2E2E}} \color[HTML]{F1F1F1} 4.11 \\
zh-en@user-review:total & {\cellcolor[HTML]{FFFDFD}} \color[HTML]{000000} 2.28 & {\cellcolor[HTML]{006400}} \color[HTML]{F1F1F1} 1.73 & {\cellcolor[HTML]{8B0000}} \color[HTML]{F1F1F1} \bfseries 2.93* & {\cellcolor[HTML]{FC1D1D}} \color[HTML]{F1F1F1} 2.71 & {\cellcolor[HTML]{DD0000}} \color[HTML]{F1F1F1} \bfseries 2.83* & {\cellcolor[HTML]{F83A3A}} \color[HTML]{F1F1F1} 2.62 & {\cellcolor[HTML]{F27171}} \color[HTML]{F1F1F1} 2.45 & {\cellcolor[HTML]{C8FDC8}} \color[HTML]{000000} 2.22 & {\cellcolor[HTML]{5AC95A}} \color[HTML]{000000} 2.06 & {\cellcolor[HTML]{F17B7B}} \color[HTML]{F1F1F1} 2.42 & {\cellcolor[HTML]{F18A8A}} \color[HTML]{F1F1F1} 2.39 & {\cellcolor[HTML]{40B440}} \color[HTML]{F1F1F1} 2.01 \\
zh-en@manuals:total & {\cellcolor[HTML]{FFFDFD}} \color[HTML]{000000} 1.70 & {\cellcolor[HTML]{006400}} \color[HTML]{F1F1F1} 1.32 & {\cellcolor[HTML]{FD1515}} \color[HTML]{F1F1F1} \bfseries 2.60* & {\cellcolor[HTML]{8B0000}} \color[HTML]{F1F1F1} 2.98 & {\cellcolor[HTML]{F64A4A}} \color[HTML]{F1F1F1} 2.28 & {\cellcolor[HTML]{F55656}} \color[HTML]{F1F1F1} 2.21 & {\cellcolor[HTML]{F17676}} \color[HTML]{F1F1F1} 2.01 & {\cellcolor[HTML]{F83F3F}} \color[HTML]{F1F1F1} 2.35 & {\cellcolor[HTML]{79E279}} \color[HTML]{000000} 1.58 & {\cellcolor[HTML]{5AC95A}} \color[HTML]{000000} 1.55 & {\cellcolor[HTML]{EF0000}} \color[HTML]{F1F1F1} 2.76 & {\cellcolor[HTML]{F49E9E}} \color[HTML]{000000} 1.89 \\

\vspace{0mm} \\

en-de@news:fluency & {\cellcolor[HTML]{FFFDFD}} \color[HTML]{000000} 0.38 & {\cellcolor[HTML]{F26F6F}} \color[HTML]{F1F1F1} 0.69 & {\cellcolor[HTML]{E10000}} \color[HTML]{F1F1F1} 1.33 & {\cellcolor[HTML]{F46060}} \color[HTML]{F1F1F1} 0.77 & {\cellcolor[HTML]{006400}} \color[HTML]{F1F1F1} 0.31 & {\cellcolor[HTML]{8B0000}} \color[HTML]{F1F1F1} 1.49 & {\cellcolor[HTML]{FBDADA}} \color[HTML]{000000} 0.44 & {\cellcolor[HTML]{F9D0D0}} \color[HTML]{000000} 0.46 & {\cellcolor[HTML]{F55151}} \color[HTML]{F1F1F1} 0.84 & {\cellcolor[HTML]{F17474}} \color[HTML]{F1F1F1} 0.66 & {\cellcolor[HTML]{79E279}} \color[HTML]{000000} 0.36 & {\cellcolor[HTML]{FCE4E4}} \color[HTML]{000000} 0.42 \\
en-de@user-review:fluency & {\cellcolor[HTML]{FFFDFD}} \color[HTML]{000000} 0.57 & {\cellcolor[HTML]{F17878}} \color[HTML]{F1F1F1} 0.65 & {\cellcolor[HTML]{36AC36}} \color[HTML]{F1F1F1} 0.37 & {\cellcolor[HTML]{F55454}} \color[HTML]{F1F1F1} 0.70 & {\cellcolor[HTML]{B8FCB8}} \color[HTML]{000000} 0.52 & {\cellcolor[HTML]{8B0000}} \color[HTML]{F1F1F1} 0.89 & {\cellcolor[HTML]{9D0000}} \color[HTML]{F1F1F1} 0.88 & {\cellcolor[HTML]{D1FDD1}} \color[HTML]{000000} 0.53 & {\cellcolor[HTML]{98FB98}} \color[HTML]{000000} 0.49 & {\cellcolor[HTML]{FC1C1C}} \color[HTML]{F1F1F1} 0.79 & {\cellcolor[HTML]{006400}} \color[HTML]{F1F1F1} 0.18 & {\cellcolor[HTML]{F8C6C6}} \color[HTML]{000000} 0.60 \\
en-de@mastodon:fluency & {\cellcolor[HTML]{FFFDFD}} \color[HTML]{000000} 0.15 & {\cellcolor[HTML]{F55858}} \color[HTML]{F1F1F1} 0.21 & {\cellcolor[HTML]{F29494}} \color[HTML]{000000} 0.17 & {\cellcolor[HTML]{40B440}} \color[HTML]{F1F1F1} 0.13 & {\cellcolor[HTML]{FC1717}} \color[HTML]{F1F1F1} 0.25 & {\cellcolor[HTML]{8B0000}} \color[HTML]{F1F1F1} \bfseries 0.30* & {\cellcolor[HTML]{CDFDCD}} \color[HTML]{000000} 0.15 & {\cellcolor[HTML]{F07D7D}} \color[HTML]{F1F1F1} 0.18 & {\cellcolor[HTML]{006400}} \color[HTML]{F1F1F1} 0.12 & {\cellcolor[HTML]{D5FDD5}} \color[HTML]{000000} 0.15 & {\cellcolor[HTML]{F74444}} \color[HTML]{F1F1F1} 0.22 & {\cellcolor[HTML]{F55A5A}} \color[HTML]{F1F1F1} 0.21 \\
en-de@speech:fluency & {\cellcolor[HTML]{FFFDFD}} \color[HTML]{000000} 1.21 & {\cellcolor[HTML]{098709}} \color[HTML]{F1F1F1} 0.63 & {\cellcolor[HTML]{007300}} \color[HTML]{F1F1F1} 0.48 & {\cellcolor[HTML]{006400}} \color[HTML]{F1F1F1} 0.34 & {\cellcolor[HTML]{A0FBA0}} \color[HTML]{000000} \bfseries 1.05* & {\cellcolor[HTML]{68D468}} \color[HTML]{000000} 0.90 & {\cellcolor[HTML]{007300}} \color[HTML]{F1F1F1} 0.47 & {\cellcolor[HTML]{93F793}} \color[HTML]{000000} 1.03 & {\cellcolor[HTML]{5CCB5C}} \color[HTML]{000000} 0.87 & {\cellcolor[HTML]{239C23}} \color[HTML]{F1F1F1} 0.70 & {\cellcolor[HTML]{007300}} \color[HTML]{F1F1F1} 0.48 & {\cellcolor[HTML]{007A00}} \color[HTML]{F1F1F1} 0.55 \\
zh-en@news:fluency & {\cellcolor[HTML]{FFFDFD}} \color[HTML]{000000} 0.29 & {\cellcolor[HTML]{8B0000}} \color[HTML]{F1F1F1} 1.02 & {\cellcolor[HTML]{F28F8F}} \color[HTML]{F1F1F1} 0.42 & {\cellcolor[HTML]{F8C1C1}} \color[HTML]{000000} 0.36 & {\cellcolor[HTML]{F6B2B2}} \color[HTML]{000000} 0.38 & {\cellcolor[HTML]{F17979}} \color[HTML]{F1F1F1} 0.46 & {\cellcolor[HTML]{E9FEE9}} \color[HTML]{000000} 0.29 & {\cellcolor[HTML]{FCE4E4}} \color[HTML]{000000} 0.32 & {\cellcolor[HTML]{F6B7B7}} \color[HTML]{000000} 0.37 & {\cellcolor[HTML]{F28F8F}} \color[HTML]{F1F1F1} 0.42 & {\cellcolor[HTML]{006400}} \color[HTML]{F1F1F1} 0.25 & {\cellcolor[HTML]{FBDFDF}} \color[HTML]{000000} 0.33 \\
zh-en@user-review:fluency & {\cellcolor[HTML]{FFFDFD}} \color[HTML]{000000} 0.51 & {\cellcolor[HTML]{007600}} \color[HTML]{F1F1F1} 0.18 & {\cellcolor[HTML]{007E00}} \color[HTML]{F1F1F1} \bfseries 0.22* & {\cellcolor[HTML]{007500}} \color[HTML]{F1F1F1} 0.18 & {\cellcolor[HTML]{048304}} \color[HTML]{F1F1F1} \bfseries 0.23* & {\cellcolor[HTML]{45B745}} \color[HTML]{F1F1F1} 0.31 & {\cellcolor[HTML]{47B947}} \color[HTML]{F1F1F1} 0.32 & {\cellcolor[HTML]{4CBD4C}} \color[HTML]{F1F1F1} 0.33 & {\cellcolor[HTML]{5AC95A}} \color[HTML]{000000} 0.34 & {\cellcolor[HTML]{6FDA6F}} \color[HTML]{000000} 0.37 & {\cellcolor[HTML]{006400}} \color[HTML]{F1F1F1} 0.10 & {\cellcolor[HTML]{007F00}} \color[HTML]{F1F1F1} 0.22 \\
zh-en@manuals:fluency & {\cellcolor[HTML]{FFFDFD}} \color[HTML]{000000} 0.20 & {\cellcolor[HTML]{F83A3A}} \color[HTML]{F1F1F1} 0.47 & {\cellcolor[HTML]{F18A8A}} \color[HTML]{F1F1F1} \bfseries 0.29* & {\cellcolor[HTML]{F36464}} \color[HTML]{F1F1F1} 0.37 & {\cellcolor[HTML]{F39999}} \color[HTML]{000000} 0.28 & {\cellcolor[HTML]{8B0000}} \color[HTML]{F1F1F1} 0.71 & {\cellcolor[HTML]{F74747}} \color[HTML]{F1F1F1} 0.43 & {\cellcolor[HTML]{FE0909}} \color[HTML]{F1F1F1} 0.58 & {\cellcolor[HTML]{FA2929}} \color[HTML]{F1F1F1} 0.51 & {\cellcolor[HTML]{F64949}} \color[HTML]{F1F1F1} 0.43 & {\cellcolor[HTML]{FD1010}} \color[HTML]{F1F1F1} 0.57 & {\cellcolor[HTML]{900000}} \color[HTML]{F1F1F1} 0.70 \\

\vspace{0mm} \\

en-de@news:accuracy & {\cellcolor[HTML]{FFFDFD}} \color[HTML]{000000} 0.65 & {\cellcolor[HTML]{FA2929}} \color[HTML]{F1F1F1} 1.55 & {\cellcolor[HTML]{FB1F1F}} \color[HTML]{F1F1F1} 1.63 & {\cellcolor[HTML]{8B0000}} \color[HTML]{F1F1F1} 2.14 & {\cellcolor[HTML]{FB2424}} \color[HTML]{F1F1F1} 1.59 & {\cellcolor[HTML]{FB1F1F}} \color[HTML]{F1F1F1} 1.63 & {\cellcolor[HTML]{F07E7E}} \color[HTML]{F1F1F1} 0.96 & {\cellcolor[HTML]{F07D7D}} \color[HTML]{F1F1F1} 0.97 & {\cellcolor[HTML]{006400}} \color[HTML]{F1F1F1} 0.59 & {\cellcolor[HTML]{F8C6C6}} \color[HTML]{000000} 0.78 & {\cellcolor[HTML]{F36969}} \color[HTML]{F1F1F1} 1.12 & {\cellcolor[HTML]{F08080}} \color[HTML]{F1F1F1} 0.95 \\
en-de@user-review:accuracy & {\cellcolor[HTML]{FFFDFD}} \color[HTML]{000000} 2.32 & {\cellcolor[HTML]{007E00}} \color[HTML]{F1F1F1} 1.25 & {\cellcolor[HTML]{006800}} \color[HTML]{F1F1F1} 0.89 & {\cellcolor[HTML]{28A028}} \color[HTML]{F1F1F1} 1.47 & {\cellcolor[HTML]{007800}} \color[HTML]{F1F1F1} 1.16 & {\cellcolor[HTML]{8CF18C}} \color[HTML]{000000} 1.96 & {\cellcolor[HTML]{68D468}} \color[HTML]{000000} 1.79 & {\cellcolor[HTML]{8B0000}} \color[HTML]{F1F1F1} 2.37 & {\cellcolor[HTML]{007E00}} \color[HTML]{F1F1F1} 1.25 & {\cellcolor[HTML]{098709}} \color[HTML]{F1F1F1} 1.32 & {\cellcolor[HTML]{006F00}} \color[HTML]{F1F1F1} 1.00 & {\cellcolor[HTML]{006400}} \color[HTML]{F1F1F1} 0.82 \\
en-de@mastodon:accuracy & {\cellcolor[HTML]{FFFDFD}} \color[HTML]{000000} 0.54 & {\cellcolor[HTML]{AB0000}} \color[HTML]{F1F1F1} 1.06 & {\cellcolor[HTML]{F27373}} \color[HTML]{F1F1F1} 0.68 & {\cellcolor[HTML]{FE0606}} \color[HTML]{F1F1F1} 0.97 & {\cellcolor[HTML]{FE0D0D}} \color[HTML]{F1F1F1} 0.95 & {\cellcolor[HTML]{8B0000}} \color[HTML]{F1F1F1} \bfseries 1.10* & {\cellcolor[HTML]{F39999}} \color[HTML]{000000} 0.63 & {\cellcolor[HTML]{FBDFDF}} \color[HTML]{000000} 0.57 & {\cellcolor[HTML]{FBDFDF}} \color[HTML]{000000} 0.57 & {\cellcolor[HTML]{D1FDD1}} \color[HTML]{000000} 0.53 & {\cellcolor[HTML]{E10000}} \color[HTML]{F1F1F1} 1.01 & {\cellcolor[HTML]{006400}} \color[HTML]{F1F1F1} 0.47 \\
en-de@speech:accuracy & {\cellcolor[HTML]{FFFDFD}} \color[HTML]{000000} 1.77 & {\cellcolor[HTML]{F46262}} \color[HTML]{F1F1F1} 2.41 & {\cellcolor[HTML]{F45F5F}} \color[HTML]{F1F1F1} 2.44 & {\cellcolor[HTML]{31A831}} \color[HTML]{F1F1F1} 1.66 & {\cellcolor[HTML]{8B0000}} \color[HTML]{F1F1F1} \bfseries 3.65* & {\cellcolor[HTML]{F18585}} \color[HTML]{F1F1F1} 2.13 & {\cellcolor[HTML]{F8C6C6}} \color[HTML]{000000} 1.94 & {\cellcolor[HTML]{006400}} \color[HTML]{F1F1F1} 1.56 & {\cellcolor[HTML]{006400}} \color[HTML]{F1F1F1} 1.56 & {\cellcolor[HTML]{007600}} \color[HTML]{F1F1F1} 1.61 & {\cellcolor[HTML]{006400}} \color[HTML]{F1F1F1} 1.56 & {\cellcolor[HTML]{31A831}} \color[HTML]{F1F1F1} 1.66 \\
zh-en@news:accuracy & {\cellcolor[HTML]{FFFDFD}} \color[HTML]{000000} 3.03 & {\cellcolor[HTML]{F5A8A8}} \color[HTML]{000000} 3.10 & {\cellcolor[HTML]{F46060}} \color[HTML]{F1F1F1} 3.21 & {\cellcolor[HTML]{FC1A1A}} \color[HTML]{F1F1F1} 3.39 & {\cellcolor[HTML]{F07D7D}} \color[HTML]{F1F1F1} 3.14 & {\cellcolor[HTML]{FB2626}} \color[HTML]{F1F1F1} 3.36 & {\cellcolor[HTML]{61CF61}} \color[HTML]{000000} 2.79 & {\cellcolor[HTML]{006500}} \color[HTML]{F1F1F1} 2.40 & {\cellcolor[HTML]{F46060}} \color[HTML]{F1F1F1} 3.21 & {\cellcolor[HTML]{006400}} \color[HTML]{F1F1F1} 2.39 & {\cellcolor[HTML]{FC1A1A}} \color[HTML]{F1F1F1} 3.39 & {\cellcolor[HTML]{8B0000}} \color[HTML]{F1F1F1} 3.56 \\
zh-en@user-review:accuracy & {\cellcolor[HTML]{FFFDFD}} \color[HTML]{000000} 1.23 & {\cellcolor[HTML]{006400}} \color[HTML]{F1F1F1} 1.21 & {\cellcolor[HTML]{8B0000}} \color[HTML]{F1F1F1} \bfseries 2.36* & {\cellcolor[HTML]{D80000}} \color[HTML]{F1F1F1} 2.20 & {\cellcolor[HTML]{C60000}} \color[HTML]{F1F1F1} \bfseries 2.24* & {\cellcolor[HTML]{FA2727}} \color[HTML]{F1F1F1} 1.92 & {\cellcolor[HTML]{F74747}} \color[HTML]{F1F1F1} 1.75 & {\cellcolor[HTML]{F17B7B}} \color[HTML]{F1F1F1} 1.48 & {\cellcolor[HTML]{F6B7B7}} \color[HTML]{000000} 1.35 & {\cellcolor[HTML]{F55A5A}} \color[HTML]{F1F1F1} 1.66 & {\cellcolor[HTML]{FC1A1A}} \color[HTML]{F1F1F1} 1.99 & {\cellcolor[HTML]{F07E7E}} \color[HTML]{F1F1F1} 1.46 \\
zh-en@manuals:accuracy & {\cellcolor[HTML]{FFFDFD}} \color[HTML]{000000} 1.38 & {\cellcolor[HTML]{006400}} \color[HTML]{F1F1F1} 0.81 & {\cellcolor[HTML]{FE0909}} \color[HTML]{F1F1F1} \bfseries 2.19* & {\cellcolor[HTML]{8B0000}} \color[HTML]{F1F1F1} 2.46 & {\cellcolor[HTML]{F74747}} \color[HTML]{F1F1F1} 1.88 & {\cellcolor[HTML]{FFFDFD}} \color[HTML]{000000} 1.38 & {\cellcolor[HTML]{F7BCBC}} \color[HTML]{000000} 1.50 & {\cellcolor[HTML]{F07D7D}} \color[HTML]{F1F1F1} 1.62 & {\cellcolor[HTML]{007D00}} \color[HTML]{F1F1F1} 0.96 & {\cellcolor[HTML]{1E991E}} \color[HTML]{F1F1F1} 1.04 & {\cellcolor[HTML]{F64F4F}} \color[HTML]{F1F1F1} 1.85 & {\cellcolor[HTML]{007000}} \color[HTML]{F1F1F1} 0.88 \\

\vspace{0mm} \\

en-de@news:other & {\cellcolor[HTML]{FFFDFD}} \color[HTML]{000000} 0.92 & {\cellcolor[HTML]{6BD66B}} \color[HTML]{000000} 0.73 & {\cellcolor[HTML]{007B00}} \color[HTML]{F1F1F1} 0.51 & {\cellcolor[HTML]{006400}} \color[HTML]{F1F1F1} 0.37 & {\cellcolor[HTML]{2AA22A}} \color[HTML]{F1F1F1} 0.62 & {\cellcolor[HTML]{2AA22A}} \color[HTML]{F1F1F1} 0.62 & {\cellcolor[HTML]{1C971C}} \color[HTML]{F1F1F1} 0.59 & {\cellcolor[HTML]{078507}} \color[HTML]{F1F1F1} 0.55 & {\cellcolor[HTML]{6BD66B}} \color[HTML]{000000} 0.73 & {\cellcolor[HTML]{007300}} \color[HTML]{F1F1F1} 0.46 & {\cellcolor[HTML]{159115}} \color[HTML]{F1F1F1} 0.58 & {\cellcolor[HTML]{006A00}} \color[HTML]{F1F1F1} 0.41 \\
en-de@user-review:other & {\cellcolor[HTML]{FFFDFD}} \color[HTML]{000000} 0.77 & {\cellcolor[HTML]{F8C1C1}} \color[HTML]{000000} 0.89 & {\cellcolor[HTML]{8B0000}} \color[HTML]{F1F1F1} 2.04 & {\cellcolor[HTML]{006400}} \color[HTML]{F1F1F1} 0.63 & {\cellcolor[HTML]{F83A3A}} \color[HTML]{F1F1F1} 1.44 & {\cellcolor[HTML]{F46060}} \color[HTML]{F1F1F1} 1.21 & {\cellcolor[HTML]{F45D5D}} \color[HTML]{F1F1F1} 1.23 & {\cellcolor[HTML]{FDEEEE}} \color[HTML]{000000} 0.81 & {\cellcolor[HTML]{F17878}} \color[HTML]{F1F1F1} 1.07 & {\cellcolor[HTML]{F18585}} \color[HTML]{F1F1F1} 1.02 & {\cellcolor[HTML]{C20000}} \color[HTML]{F1F1F1} 1.91 & {\cellcolor[HTML]{F55858}} \color[HTML]{F1F1F1} 1.26 \\
en-de@mastodon:other & {\cellcolor[HTML]{FFFDFD}} \color[HTML]{000000} 0.60 & {\cellcolor[HTML]{28A028}} \color[HTML]{F1F1F1} 0.43 & {\cellcolor[HTML]{006500}} \color[HTML]{F1F1F1} 0.31 & {\cellcolor[HTML]{64D164}} \color[HTML]{000000} 0.49 & {\cellcolor[HTML]{007F00}} \color[HTML]{F1F1F1} 0.39 & {\cellcolor[HTML]{50C150}} \color[HTML]{F1F1F1} \bfseries 0.47* & {\cellcolor[HTML]{007600}} \color[HTML]{F1F1F1} 0.36 & {\cellcolor[HTML]{34AA34}} \color[HTML]{F1F1F1} 0.44 & {\cellcolor[HTML]{007400}} \color[HTML]{F1F1F1} 0.36 & {\cellcolor[HTML]{007300}} \color[HTML]{F1F1F1} 0.35 & {\cellcolor[HTML]{1C971C}} \color[HTML]{F1F1F1} 0.42 & {\cellcolor[HTML]{006400}} \color[HTML]{F1F1F1} 0.30 \\
en-de@speech:other & {\cellcolor[HTML]{FFFDFD}} \color[HTML]{000000} 0.61 & {\cellcolor[HTML]{FD1212}} \color[HTML]{F1F1F1} 0.75 & {\cellcolor[HTML]{49BB49}} \color[HTML]{F1F1F1} 0.45 & {\cellcolor[HTML]{FFFDFD}} \color[HTML]{000000} 0.61 & {\cellcolor[HTML]{FB2121}} \color[HTML]{F1F1F1} \bfseries 0.73* & {\cellcolor[HTML]{8B0000}} \color[HTML]{F1F1F1} 0.80 & {\cellcolor[HTML]{A8FCA8}} \color[HTML]{000000} 0.55 & {\cellcolor[HTML]{098709}} \color[HTML]{F1F1F1} 0.38 & {\cellcolor[HTML]{49BB49}} \color[HTML]{F1F1F1} 0.45 & {\cellcolor[HTML]{007700}} \color[HTML]{F1F1F1} 0.34 & {\cellcolor[HTML]{C0FDC0}} \color[HTML]{000000} 0.56 & {\cellcolor[HTML]{006400}} \color[HTML]{F1F1F1} 0.27 \\
zh-en@news:other & {\cellcolor[HTML]{FFFDFD}} \color[HTML]{000000} 0.24 & {\cellcolor[HTML]{8B0000}} \color[HTML]{F1F1F1} 0.39 & {\cellcolor[HTML]{F08080}} \color[HTML]{F1F1F1} 0.27 & {\cellcolor[HTML]{006400}} \color[HTML]{F1F1F1} 0.16 & {\cellcolor[HTML]{F83F3F}} \color[HTML]{F1F1F1} 0.31 & {\cellcolor[HTML]{FD1515}} \color[HTML]{F1F1F1} 0.34 & {\cellcolor[HTML]{F36A6A}} \color[HTML]{F1F1F1} 0.29 & {\cellcolor[HTML]{F8C1C1}} \color[HTML]{000000} 0.26 & {\cellcolor[HTML]{F83F3F}} \color[HTML]{F1F1F1} 0.31 & {\cellcolor[HTML]{FD1515}} \color[HTML]{F1F1F1} 0.34 & {\cellcolor[HTML]{098709}} \color[HTML]{F1F1F1} 0.19 & {\cellcolor[HTML]{A8FCA8}} \color[HTML]{000000} 0.23 \\
zh-en@user-review:other & {\cellcolor[HTML]{FFFDFD}} \color[HTML]{000000} 0.54 & {\cellcolor[HTML]{007300}} \color[HTML]{F1F1F1} 0.34 & {\cellcolor[HTML]{007600}} \color[HTML]{F1F1F1} \bfseries 0.35* & {\cellcolor[HTML]{006F00}} \color[HTML]{F1F1F1} 0.33 & {\cellcolor[HTML]{007D00}} \color[HTML]{F1F1F1} \bfseries 0.36* & {\cellcolor[HTML]{0B890B}} \color[HTML]{F1F1F1} 0.38 & {\cellcolor[HTML]{048304}} \color[HTML]{F1F1F1} 0.38 & {\cellcolor[HTML]{31A831}} \color[HTML]{F1F1F1} 0.41 & {\cellcolor[HTML]{007B00}} \color[HTML]{F1F1F1} 0.36 & {\cellcolor[HTML]{128F12}} \color[HTML]{F1F1F1} 0.39 & {\cellcolor[HTML]{006400}} \color[HTML]{F1F1F1} 0.30 & {\cellcolor[HTML]{006D00}} \color[HTML]{F1F1F1} 0.32 \\
zh-en@manuals:other & {\cellcolor[HTML]{FFFDFD}} \color[HTML]{000000} 0.12 & {\cellcolor[HTML]{006400}} \color[HTML]{F1F1F1} 0.04 & {\cellcolor[HTML]{FFFDFD}} \color[HTML]{000000} \bfseries 0.12* & {\cellcolor[HTML]{F29494}} \color[HTML]{000000} 0.15 & {\cellcolor[HTML]{FFFDFD}} \color[HTML]{000000} 0.12 & {\cellcolor[HTML]{FFFDFD}} \color[HTML]{000000} 0.12 & {\cellcolor[HTML]{3DB23D}} \color[HTML]{F1F1F1} 0.08 & {\cellcolor[HTML]{F29494}} \color[HTML]{000000} 0.15 & {\cellcolor[HTML]{FFFDFD}} \color[HTML]{000000} 0.12 & {\cellcolor[HTML]{3DB23D}} \color[HTML]{F1F1F1} 0.08 & {\cellcolor[HTML]{8B0000}} \color[HTML]{F1F1F1} 0.35 & {\cellcolor[HTML]{EB0000}} \color[HTML]{F1F1F1} 0.31 \\
\end{tabular}

\end{small}
\caption{Human evaluation results broken down by domain and MQM error type for en-de and zh-en. Columns indicate the system used for MBR/QE decoding; ensembles are defined in Table~\ref{tab:ensembles}. 1\textsuperscript{st} block is total error scores, 2\textsuperscript{nd} is fluency error scores, 3\textsuperscript{rd} is accuracy error scores, 4\textsuperscript{th} is other error scores. For each system, average human evaluation scores across the evaluated segments are shown. Lower scores are better. Colors are relative to greedy, green is better than greedy, red is worse. Significant differences from greedy (pairwise t-test) indicated by * for p<0.05, † for p<0.01, ‡ for p<0.001.}
\label{tab:human_eval_domains}

\end{table*}

\subsection{Results}

Results are shown in Table~\ref{tab:human_eval_error_types}. We observe that overall the best-performing system is rankAvg:noNC, which significantly outperforms greedy (p<0.001 on pairwise t-test). rankAvg:noNC also performs the best on each language pair except en-hi. Interestingly, rankAvg:noNC and greedy decoding beat the reference translation in all language pairs, suggesting either that the reference translations in WMT2023 and FLORES200 are of poor quality, or that Gemini's translation quality has achieved human parity for these language pairs.



A surprising result from our human evaluation was that although MBR decoding with an ensembles of metrics was judged as having superior quality to greedy decoding, MBR/QE decoding with a single metric (MetricX, MetricX-QE, XCOMET-XXL, CometKiwi23-XXL, COMET22, AfriCOMET, AfriCOMET-QE, IndicCOMET) did not generally improve over greedy decoding (Table~\ref{tab:human_eval_error_types}). In fact, translations from MetricX MBR decoding for zh-en, MetricX-QE decoding for en-ml, AfriCOMET-QE decoding for en-sw, and IndicCOMET MBR decoding for en-ml were rated by humans as significantly worse than greedy decoding (Table~\ref{tab:human_eval_error_types}), even though automatic evaluation with other neural metrics such as MetricX and XCOMET-XXL estimated those translations as being significantly better than greedy (Appendix~\ref{sec:perlangresults}). This suggests that evaluation with neutral metrics overestimates the quality of MBR/QE decoding, even if different metrics are used for decoding and evaluation. Our findings contrast with previous studies which find that MBR decoding with a single metric outperforms greedy decoding in human evaluations~\citep{freitag-etal-2022-high, freitag-etal-2023-epsilon, tomani2023quality}.

We hypothesize a few potential causes of the failure of single-metric MBR/QE decoding to outperform greedy decoding: firstly, machine translation quality has improved considerably in recent years. This is reflected by how in our study the greedy decoding outputs achieved better human evaluation results compared to the references generated by professional human translators, especially when looking at fluency scores (Table~\ref{tab:human_eval_error_types}), in contrast with previous work where reference translations were rated as better~\citep{freitag-etal-2022-high, freitag-etal-2023-epsilon}. Therefore, it is possible that improvements in greedy translation quality have reduced the quality gains from MBR/QE decoding, and have resulted in the adverse effects of metric bias from MBR/QE decoding with a single utility metric outweighing the benefits to translation quality. For example, in Table~\ref{tab:human_eval_error_types} we can see that single-metric MBR/QE decoding generally improves fluency on high-resource languages, and reduces errors in style, terminology, and locale convention (labeled ``other''). However, accuracy suffers with single-metric MBR/QE decoding for most language pairs (Table~\ref{tab:human_eval_error_types}). We show an example in Table~\ref{tab:example_translations}, where MetricX and XCOMET-XXL MBR decoding favor a fluent yet inaccurate translation. Perhaps part of the reason for this decrease in accuracy is that MBR decoding with metrics such as MetricX considers only similarity to the pseudoreferences and does not consider the source sentence, so fluent hallucinations that occur in a large number of pseudoreferences will be favored by MBR decoding. Therefore, we hypothesize that past gains from single-metric MBR/QE decoding might have been driven by improvements in fluency and style, but modern LLMs have become good at producing fluent outputs (as indicated by the low fluency error scores for the greedy condition in Table~\ref{tab:human_eval_error_types}), so we are no longer seeing overall quality improvements from single-metric MQM/QE decoding.

We also considered the effects of domain on the quality of single-metric MBR/QE decoding. Since the WMT2023 datasets which were used include novel domains such as speech transcripts and mastodon posts which are not well-represented in the data that metrics such as MetricX and XCOMET-XXL were trained on, we hypothesized that this may adversely impacting MBR quality. However, contrary to our expectations, as we can observe in Table~\ref{tab:human_eval_domains} there is no clear effect of the domain on the quality of MBR decoding results. Thus, we do not believe effects of domain to be the primary factor behind our findings.

We also considered whether MBR decoding with other metrics we did not evaluate with human raters, such as BLEURT, would have performed better than the metrics we evaluated. To do so, we looked at the correlation between the MQM scores from our human evaluation, compared to the scores assigned by metrics. We include scores from QE metrics (to simulate QE decoding), scores from reference-based metrics based on the 128 pseudoreferences (to simulate MBR decoding), as well as scores form reference-based metrics using the actual references (to simulate a reference-based metric oracle). Table~\ref{tab:metric_correlation_kendalltau} shows Kendall-Tau correlation and Table~\ref{tab:metric_correlation_pearson} shows Pearson correlation. Note that this an imperfect simulation of what would happen if we actually performed human evaluation with the MBR/QE decoding outputs for these metrics, as we are considering correlations with human judgements only the subset of candidates which were evaluated (which is a biased sample, as they are the results of MBR/QE decoding), not all 128 samples. We observe that among the individual metrics that we did not evaluate, simulated XCOMET-XL MBR decoding seems to correlate the best with human judgements, and the other metrics are generally worse than MetricX/XCOMET-XXL MBR decoding. We also include some ensembles, finding that they are generally better correlated with human judgements than individual metrics in our simulation. Therefore, we do not expect that changing to another metric for MBR/QE decoding would have resulted in significantly better translation quality.

\section{Discussion}

While previous work has sometimes assumed that MBR decoding outputs can be evaluated by automated metrics so long as a non-utility metric is used~\citep{tomani2023quality}, we find MBR/QE decoding outputs are often preferred by automated metrics despite the fact that human raters believe they are worse quality. For example, while MetricX-QE decoding outputs are considered by human raters to be of worse quality than greedy decoding (Table~\ref{tab:human_eval_error_types}), they still achieve higher scores when evaluated by XCOMET-XXL, XCOMET-XL, MetricX, CometKiwi22, CometKiwi23-XL, and CometKiwi23-XXL (Table~\ref{tab:subset_all_test} and Appendix~\ref{sec:perlangresults}). Thus, the metric bias issue that results from MBR/QE decoding complicates evaluation with automated metrics.

\begin{table}[!tb]
\begin{small}
\setlength\tabcolsep{2pt}
\begin{tabular}{lrrrrrr}
& zh-en & en-de & en-ha & en-sw & en-hi & en-ml \\
XCOMET-XXL & {\cellcolor[HTML]{359A35}} \color[HTML]{F1F1F1} 0.278 & {\cellcolor[HTML]{44A144}} \color[HTML]{F1F1F1} 0.110 & {\cellcolor[HTML]{329832}} \color[HTML]{F1F1F1} 0.114 & {\cellcolor[HTML]{219021}} \color[HTML]{F1F1F1} 0.201 & {\cellcolor[HTML]{91C791}} \color[HTML]{000000} 0.073 & {\cellcolor[HTML]{57AB57}} \color[HTML]{F1F1F1} 0.152 \\
\hline
\scriptsize XCOMET-XXL:mbr & {\cellcolor[HTML]{389B38}} \color[HTML]{F1F1F1} 0.275 & {\cellcolor[HTML]{42A042}} \color[HTML]{F1F1F1} 0.111 & {\cellcolor[HTML]{1B8D1B}} \color[HTML]{F1F1F1} 0.125 & {\cellcolor[HTML]{0F870F}} \color[HTML]{F1F1F1} 0.212 & {\cellcolor[HTML]{379B37}} \color[HTML]{F1F1F1} 0.094 & {\cellcolor[HTML]{57AB57}} \color[HTML]{F1F1F1} 0.152 \\
\hline
XCOMET-XL & {\cellcolor[HTML]{098509}} \color[HTML]{F1F1F1} 0.335 & {\cellcolor[HTML]{1F8F1F}} \color[HTML]{F1F1F1} 0.126 & {\cellcolor[HTML]{1F8F1F}} \color[HTML]{F1F1F1} 0.123 & {\cellcolor[HTML]{399C39}} \color[HTML]{F1F1F1} 0.187 & {\cellcolor[HTML]{58AB58}} \color[HTML]{F1F1F1} 0.087 & {\cellcolor[HTML]{1E8F1E}} \color[HTML]{F1F1F1} 0.179 \\
\hline
\scriptsize XCOMET-XL:mbr & {\cellcolor[HTML]{098509}} \color[HTML]{F1F1F1} 0.336 & {\cellcolor[HTML]{0D860D}} \color[HTML]{F1F1F1} 0.134 & {\cellcolor[HTML]{008000}} \color[HTML]{F1F1F1} 0.137 & {\cellcolor[HTML]{229122}} \color[HTML]{F1F1F1} 0.201 & {\cellcolor[HTML]{3D9E3D}} \color[HTML]{F1F1F1} 0.093 & {\cellcolor[HTML]{359A35}} \color[HTML]{F1F1F1} 0.168 \\
\hline
MetricX & {\cellcolor[HTML]{4AA44A}} \color[HTML]{F1F1F1} 0.252 & {\cellcolor[HTML]{B0D6B0}} \color[HTML]{000000} 0.065 & {\cellcolor[HTML]{86C286}} \color[HTML]{000000} 0.077 & {\cellcolor[HTML]{319831}} \color[HTML]{F1F1F1} 0.192 & {\cellcolor[HTML]{57AB57}} \color[HTML]{F1F1F1} 0.087 & {\cellcolor[HTML]{54A954}} \color[HTML]{F1F1F1} 0.154 \\
\hline
MetricX:mbr & {\cellcolor[HTML]{2D962D}} \color[HTML]{F1F1F1} 0.289 & {\cellcolor[HTML]{78BB78}} \color[HTML]{F1F1F1} 0.089 & {\cellcolor[HTML]{3A9C3A}} \color[HTML]{F1F1F1} 0.111 & {\cellcolor[HTML]{118811}} \color[HTML]{F1F1F1} 0.211 & {\cellcolor[HTML]{2B952B}} \color[HTML]{F1F1F1} 0.097 & {\cellcolor[HTML]{5FAE5F}} \color[HTML]{F1F1F1} 0.149 \\
\hline
MetricX-QE & {\cellcolor[HTML]{2C962C}} \color[HTML]{F1F1F1} 0.291 & {\cellcolor[HTML]{DBEBDB}} \color[HTML]{000000} 0.046 & {\cellcolor[HTML]{61B061}} \color[HTML]{F1F1F1} 0.093 & {\cellcolor[HTML]{5FAE5F}} \color[HTML]{F1F1F1} 0.166 & {\cellcolor[HTML]{B4D8B4}} \color[HTML]{000000} 0.065 & {\cellcolor[HTML]{86C286}} \color[HTML]{000000} 0.130 \\
\hline
\scriptsize CometKiwi23-XXL & {\cellcolor[HTML]{40A040}} \color[HTML]{F1F1F1} 0.264 & {\cellcolor[HTML]{8CC58C}} \color[HTML]{000000} 0.080 & {\cellcolor[HTML]{309730}} \color[HTML]{F1F1F1} 0.115 & {\cellcolor[HTML]{69B369}} \color[HTML]{F1F1F1} 0.160 & {\cellcolor[HTML]{60AF60}} \color[HTML]{F1F1F1} 0.085 & {\cellcolor[HTML]{71B771}} \color[HTML]{F1F1F1} 0.140 \\
\hline
\scriptsize CometKiwi23-XL & {\cellcolor[HTML]{349934}} \color[HTML]{F1F1F1} 0.281 & {\cellcolor[HTML]{6BB46B}} \color[HTML]{F1F1F1} 0.094 & {\cellcolor[HTML]{359A35}} \color[HTML]{F1F1F1} 0.113 & {\cellcolor[HTML]{8FC68F}} \color[HTML]{000000} 0.138 & {\cellcolor[HTML]{1D8E1D}} \color[HTML]{F1F1F1} 0.101 & {\cellcolor[HTML]{3C9D3C}} \color[HTML]{F1F1F1} 0.165 \\
\hline
CometKiwi22 & {\cellcolor[HTML]{399C39}} \color[HTML]{F1F1F1} 0.274 & {\cellcolor[HTML]{4CA54C}} \color[HTML]{F1F1F1} 0.107 & {\cellcolor[HTML]{EBF3EB}} \color[HTML]{000000} 0.032 & {\cellcolor[HTML]{52A852}} \color[HTML]{F1F1F1} 0.173 & {\cellcolor[HTML]{56AA56}} \color[HTML]{F1F1F1} 0.087 & {\cellcolor[HTML]{1E8F1E}} \color[HTML]{F1F1F1} 0.179 \\
\hline
COMET22 & {\cellcolor[HTML]{3C9D3C}} \color[HTML]{F1F1F1} 0.271 & {\cellcolor[HTML]{5CAD5C}} \color[HTML]{F1F1F1} 0.100 & {\cellcolor[HTML]{A7D2A7}} \color[HTML]{000000} 0.062 & {\cellcolor[HTML]{49A449}} \color[HTML]{F1F1F1} 0.179 & {\cellcolor[HTML]{83C083}} \color[HTML]{000000} 0.076 & {\cellcolor[HTML]{399C39}} \color[HTML]{F1F1F1} 0.166 \\
\hline
COMET22:mbr & {\cellcolor[HTML]{2C962C}} \color[HTML]{F1F1F1} 0.290 & {\cellcolor[HTML]{239123}} \color[HTML]{F1F1F1} 0.125 & {\cellcolor[HTML]{9DCD9D}} \color[HTML]{000000} 0.067 & {\cellcolor[HTML]{41A041}} \color[HTML]{F1F1F1} 0.183 & {\cellcolor[HTML]{54A954}} \color[HTML]{F1F1F1} 0.088 & {\cellcolor[HTML]{49A449}} \color[HTML]{F1F1F1} 0.159 \\
\hline
BLEURT & {\cellcolor[HTML]{359A35}} \color[HTML]{F1F1F1} 0.279 & {\cellcolor[HTML]{1B8D1B}} \color[HTML]{F1F1F1} 0.128 & {\cellcolor[HTML]{56AA56}} \color[HTML]{F1F1F1} 0.098 & {\cellcolor[HTML]{52A852}} \color[HTML]{F1F1F1} 0.173 & {\cellcolor[HTML]{66B266}} \color[HTML]{F1F1F1} 0.083 & {\cellcolor[HTML]{65B265}} \color[HTML]{F1F1F1} 0.146 \\
\hline
BLEURT:mbr & {\cellcolor[HTML]{3C9D3C}} \color[HTML]{F1F1F1} 0.271 & {\cellcolor[HTML]{0D860D}} \color[HTML]{F1F1F1} 0.134 & {\cellcolor[HTML]{279327}} \color[HTML]{F1F1F1} 0.119 & {\cellcolor[HTML]{3B9D3B}} \color[HTML]{F1F1F1} 0.187 & {\cellcolor[HTML]{008000}} \color[HTML]{F1F1F1} 0.108 & {\cellcolor[HTML]{84C084}} \color[HTML]{000000} 0.132 \\
\hline
YiSi & {\cellcolor[HTML]{84C084}} \color[HTML]{000000} 0.178 & {\cellcolor[HTML]{D4E8D4}} \color[HTML]{000000} 0.049 & {\cellcolor[HTML]{91C791}} \color[HTML]{000000} 0.072 & {\cellcolor[HTML]{CBE4CB}} \color[HTML]{000000} 0.105 & {\cellcolor[HTML]{C3E0C3}} \color[HTML]{000000} 0.061 & {\cellcolor[HTML]{75B975}} \color[HTML]{F1F1F1} 0.138 \\
\hline
YiSi:mbr & {\cellcolor[HTML]{7FBE7F}} \color[HTML]{000000} 0.183 & {\cellcolor[HTML]{A9D3A9}} \color[HTML]{000000} 0.068 & {\cellcolor[HTML]{5BAD5B}} \color[HTML]{F1F1F1} 0.096 & {\cellcolor[HTML]{B1D7B1}} \color[HTML]{000000} 0.119 & {\cellcolor[HTML]{B3D8B3}} \color[HTML]{000000} 0.065 & {\cellcolor[HTML]{55A955}} \color[HTML]{F1F1F1} 0.154 \\
\hline
chrF & {\cellcolor[HTML]{EBF3EB}} \color[HTML]{000000} 0.044 & {\cellcolor[HTML]{EBF3EB}} \color[HTML]{000000} 0.040 & {\cellcolor[HTML]{78BB78}} \color[HTML]{F1F1F1} 0.083 & {\cellcolor[HTML]{B9DBB9}} \color[HTML]{000000} 0.115 & {\cellcolor[HTML]{A8D2A8}} \color[HTML]{000000} 0.067 & {\cellcolor[HTML]{8AC48A}} \color[HTML]{000000} 0.129 \\
\hline
chrF:mbr & {\cellcolor[HTML]{C7E1C7}} \color[HTML]{000000} 0.091 & {\cellcolor[HTML]{D4E8D4}} \color[HTML]{000000} 0.049 & {\cellcolor[HTML]{56AA56}} \color[HTML]{F1F1F1} 0.098 & {\cellcolor[HTML]{96C996}} \color[HTML]{000000} 0.135 & {\cellcolor[HTML]{DAEBDA}} \color[HTML]{000000} 0.056 & {\cellcolor[HTML]{66B266}} \color[HTML]{F1F1F1} 0.146 \\
\hline
chrF++ & {\cellcolor[HTML]{E0EEE0}} \color[HTML]{000000} 0.057 & {\cellcolor[HTML]{DFEDDF}} \color[HTML]{000000} 0.045 & {\cellcolor[HTML]{77BA77}} \color[HTML]{F1F1F1} 0.084 & {\cellcolor[HTML]{B3D8B3}} \color[HTML]{000000} 0.118 & {\cellcolor[HTML]{B7DAB7}} \color[HTML]{000000} 0.064 & {\cellcolor[HTML]{97CA97}} \color[HTML]{000000} 0.123 \\
\hline
chrF++:mbr & {\cellcolor[HTML]{BDDDBD}} \color[HTML]{000000} 0.103 & {\cellcolor[HTML]{CFE5CF}} \color[HTML]{000000} 0.052 & {\cellcolor[HTML]{56AA56}} \color[HTML]{F1F1F1} 0.098 & {\cellcolor[HTML]{96C996}} \color[HTML]{000000} 0.135 & {\cellcolor[HTML]{D4E8D4}} \color[HTML]{000000} 0.057 & {\cellcolor[HTML]{6FB76F}} \color[HTML]{F1F1F1} 0.141 \\
\hline
sentBLEU & {\cellcolor[HTML]{BEDDBE}} \color[HTML]{000000} 0.102 & {\cellcolor[HTML]{BDDDBD}} \color[HTML]{000000} 0.059 & {\cellcolor[HTML]{90C790}} \color[HTML]{000000} 0.072 & {\cellcolor[HTML]{C9E2C8}} \color[HTML]{000000} 0.106 & {\cellcolor[HTML]{EBF3EB}} \color[HTML]{000000} 0.052 & {\cellcolor[HTML]{EBF3EB}} \color[HTML]{000000} 0.083 \\
\hline
sentBLEU:mbr & {\cellcolor[HTML]{95C995}} \color[HTML]{000000} 0.155 & {\cellcolor[HTML]{BFDEBF}} \color[HTML]{000000} 0.058 & {\cellcolor[HTML]{7BBC7B}} \color[HTML]{000000} 0.082 & {\cellcolor[HTML]{AED5AE}} \color[HTML]{000000} 0.121 & {\cellcolor[HTML]{D2E7D2}} \color[HTML]{000000} 0.058 & {\cellcolor[HTML]{C0DEC0}} \color[HTML]{000000} 0.103 \\
\hline
TER & {\cellcolor[HTML]{A9D3A9}} \color[HTML]{000000} 0.129 & {\cellcolor[HTML]{BADBBA}} \color[HTML]{000000} 0.061 & {\cellcolor[HTML]{76BA76}} \color[HTML]{F1F1F1} 0.084 & {\cellcolor[HTML]{EBF3EB}} \color[HTML]{000000} 0.086 & {\cellcolor[HTML]{7FBE7F}} \color[HTML]{000000} 0.077 & {\cellcolor[HTML]{E4F0E4}} \color[HTML]{000000} 0.087 \\
\hline
TER:mbr & {\cellcolor[HTML]{B5D9B5}} \color[HTML]{000000} 0.114 & {\cellcolor[HTML]{BBDCBB}} \color[HTML]{000000} 0.060 & {\cellcolor[HTML]{6DB66D}} \color[HTML]{F1F1F1} 0.088 & {\cellcolor[HTML]{D7E9D7}} \color[HTML]{000000} 0.097 & {\cellcolor[HTML]{A9D3A9}} \color[HTML]{000000} 0.067 & {\cellcolor[HTML]{A5D1A5}} \color[HTML]{000000} 0.116 \\
\hline
\vtop{\hbox{\strut MetricX}\hbox{\strut +MetricX-QE}}  & {\cellcolor[HTML]{2F972F}} \color[HTML]{F1F1F1} 0.287 & {\cellcolor[HTML]{C8E2C8}} \color[HTML]{000000} 0.055 & {\cellcolor[HTML]{76BA76}} \color[HTML]{F1F1F1} 0.084 & {\cellcolor[HTML]{2A952A}} \color[HTML]{F1F1F1} 0.196 & {\cellcolor[HTML]{54A954}} \color[HTML]{F1F1F1} 0.088 & {\cellcolor[HTML]{51A851}} \color[HTML]{F1F1F1} 0.155 \\
\hline
\vtop{\hbox{\strut MetricX}\hbox{\strut +MetricX-QE}} & {\cellcolor[HTML]{219021}} \color[HTML]{F1F1F1} 0.304 & {\cellcolor[HTML]{A4D0A4}} \color[HTML]{000000} 0.070 & {\cellcolor[HTML]{42A042}} \color[HTML]{F1F1F1} 0.107 & {\cellcolor[HTML]{1D8E1D}} \color[HTML]{F1F1F1} 0.203 & {\cellcolor[HTML]{2C962C}} \color[HTML]{F1F1F1} 0.097 & {\cellcolor[HTML]{5BAD5B}} \color[HTML]{F1F1F1} 0.151 \\
\hline
\vtop{\hbox{\strut XCOMET-XXL}\hbox{\strut +XCOMET-XL}} & {\cellcolor[HTML]{118911}} \color[HTML]{F1F1F1} 0.326 & {\cellcolor[HTML]{249224}} \color[HTML]{F1F1F1} 0.124 & {\cellcolor[HTML]{239123}} \color[HTML]{F1F1F1} 0.121 & {\cellcolor[HTML]{118911}} \color[HTML]{F1F1F1} 0.210 & {\cellcolor[HTML]{51A851}} \color[HTML]{F1F1F1} 0.088 & {\cellcolor[HTML]{0F870F}} \color[HTML]{F1F1F1} 0.186 \\
\hline
\vtop{\hbox{\strut \scriptsize XCOMET-XXL:mbr}\hbox{\strut \scriptsize +XCOMET-XL:mbr}} & {\cellcolor[HTML]{128912}} \color[HTML]{F1F1F1} 0.324 & {\cellcolor[HTML]{138913}} \color[HTML]{F1F1F1} 0.131 & {\cellcolor[HTML]{018001}} \color[HTML]{F1F1F1} 0.136 & {\cellcolor[HTML]{068306}} \color[HTML]{F1F1F1} 0.216 & {\cellcolor[HTML]{299429}} \color[HTML]{F1F1F1} 0.098 & {\cellcolor[HTML]{229122}} \color[HTML]{F1F1F1} 0.177 \\
\hline
\vtop{\hbox{\strut XCOMET-XXL}\hbox{\strut +XCOMET-XL}\hbox{\strut +COMET22}} & {\cellcolor[HTML]{028102}} \color[HTML]{F1F1F1} 0.346 & {\cellcolor[HTML]{1D8E1D}} \color[HTML]{F1F1F1} 0.127 & {\cellcolor[HTML]{2D962D}} \color[HTML]{F1F1F1} 0.116 & {\cellcolor[HTML]{0D860D}} \color[HTML]{F1F1F1} 0.213 & {\cellcolor[HTML]{4AA44A}} \color[HTML]{F1F1F1} 0.090 & {\cellcolor[HTML]{008000}} \color[HTML]{F1F1F1} 0.193 \\
\hline
\vtop{\hbox{\strut \scriptsize XCOMET-XXL:mbr}\hbox{\strut \scriptsize +XCOMET-XL:mbr}\hbox{\strut +COMET22:mbr}} & {\cellcolor[HTML]{008000}} \color[HTML]{F1F1F1} 0.348 & {\cellcolor[HTML]{008000}} \color[HTML]{F1F1F1} 0.140 & {\cellcolor[HTML]{108810}} \color[HTML]{F1F1F1} 0.129 & {\cellcolor[HTML]{008000}} \color[HTML]{F1F1F1} 0.220 & {\cellcolor[HTML]{229122}} \color[HTML]{F1F1F1} 0.100 & {\cellcolor[HTML]{148A14}} \color[HTML]{F1F1F1} 0.184 \\
\end{tabular}
\end{small}
\caption{Kendall-Tau correlation between MQM evaluation scores and automated evaluation scores. For reference-based metrics, rows with ``:mbr'' indicate pseudoreference-based evaluation. Bottom rows are ensembles that take the average between the listed metrics. Higher scores indicate better agreement with human raters. See Table~\ref{tab:metric_correlation_pearson} for Pearson correlation.}
\label{tab:metric_correlation_kendalltau}
\end{table}

That said, while we have shown that MBR/QE decoding generated translations with higher automated evaluation scores are not always judged as having better quality by humans, this does not mean that automated metrics are no longer useful. In our study, automatic reference-based metrics, QE metrics, and ensembles of metrics are still somewhat correlated with MQM scores, as shown in Table~\ref{tab:metric_correlation_kendalltau}. 
Therefore, while it is advisable to perform a human evaluation when feasible if evaluating systems that make use of MBR/QE decoding, existing metrics still correlate with human preferences. Additionally, using an ensemble of metrics for MBR decoding results in improved translation quality compared to greedy decoding and MBR/QE decoding with a single metric (Table~\ref{tab:human_eval_error_types}).

Why is it that using an ensemble of metrics for MBR decoding improves translation quality compared to just using a single metric (Table~\ref{tab:human_eval_error_types})? We hypothesize that each metric has its own biases that lead it to prefer bad translations, but different metrics have different biases, so using an ensemble reduces metric bias. We see an example of this in Table~\ref{tab:example_translations} where MetricX and XCOMET-XXL assign high scores to an inaccurate translation, but this translation is rated poorly by CometKiwi23-XXL and COMET22, so the ensemble ends up picking a good translation that is preferred by all metrics.

Techniques other than MBR/QE decoding for making use of human preferences to improve translation quality, such as DPO (direct preference optimization)~\citep{rafailov2024direct, yang-etal-2024-direct}) and RLHF (reinforcement learning from human feedback)~\citep{christiano2017deep}, might be more resilient to this metric bias issue, as they do not directly make use of the evaluation metric. However, given that the data used for DPO/RLHF is similar to the data used to train evaluation metrics, and given that the reward hacking issue is prevalent throughout reinforcement learning~\citep{skalse2022defining}, issues similar to metric bias may still occur with these techniques.

An open question that remains is how to develop new evaluation techniques that are resilient to metric bias in MBR/QE decoding. One potential way is to develop metrics specialized for evaluating MBR/QE decoding outputs from a particular system, by generating MBR/QE decoding outputs from a translation model, obtaining human annotations for those, and training a metric with them. This process is unfortunately costly and time-intensive, and the learned metric might not be able to generalize beyond translations generated by the particular utility metric and translation model it was trained on. Perhaps a better approach would be to view the metric bias problem as an adversarial learning problem, and apply techniques such as generative adversarial training~\citep{yang-etal-2018-improving} to help train metrics resilient to MBR bias.

\section{Conclusion}

In this paper we have explored the problem of metric bias, where MBR or QE decoding with a single utility metric shows improvements on automated evaluation with the utility metric and related metrics, but does not actually improve quality when judged by a human rater. We find that the metric bias issue is most severe when using a single utility metric, and using an ensemble of metrics to perform MBR decoding can help improve quality as judged by a human rater. While we have shown that metric bias can result in overly-optimistic automatic evaluations of systems that make use of MBR/QE decoding, the question of how to resolve this issue and automatically evaluate systems that make use of MBR/QE decoding is still an open problem which we leave to future work. 

\section*{Dataset}

Dataset is at \url{https://mbrbias.github.io/}


\section*{Limitations}
In this work we compare to only full MBR decoding and QE filtering as baselines, but there are many alternative approaches, such as MBR approximation heuristics~\citep{trabelsi2024efficient, jinnai2024hyperparameter, deguchi2024centroid, deguchi-etal-2023-naist, vamvas2024linear, eikema-aziz-2022-sampling}, direct preference optimization training~\citep{yang-etal-2024-direct}, quality-aware training~\citep{tomani2023quality}, or training on MBR decoding outputs ~\citep{finkelstein2023mbr}, that are more practical to use if translation latency is important. In this work we only look at translations coming from Gemini 1.0 Pro with 5-shot sample prompts and epsilon sampling, and it is possible that results may differ if using a different translation system, different prompts, or a different sampling technique. In this work we only look at using 128 samples due to the computationally expensive $O(n^2)$ cost of running full MBR decoding, but it is possible that using additional samples can achieve further quality improvements. In this work we only looked at segment-level translation, and it is possible that results may differ if performing document-level translation. However, MetricX and the COMET families of models have input token limits -- 1024 tokens for MetricX, 512 tokens for COMET -- which make it difficult to use them for document-level MBR decoding. Our human evaluation used only a single rater for each translation, which introduces the question of how reliable and consistent the ratings are -- using multiple raters and looking at inter-rater agreement is preferable, but was beyond our budget constraints. 

\section*{Ethics Statement}

MBR decoding is resource-intensive, and using ensembles of multiple metrics increases computational complexity compared to a single utility metric. To mitigate this issue, we presented two-step ensembles that use QE filtering followed by MBR decoding, which reduce the computational cost below the cost of standard MBR decoding with a single metric. 




\bibliography{bibliography,custom}
\bibliographystyle{acl_natbib}

\appendix


\section{Methodology Details}
\label{sec:methodologydetails}
\subsection{Prompts Used for Generating Samples}
\label{sec:prompts}
For each language pair, we obtained 5-shot examples for our prompts from the dev split of FLORES-200 by randomly sampling among those reference pairs that had perfect MetricX QE scores (scores of 0). We used MetricX QE filtering to ensure we used high-quality examples as our 5-shot examples. The sampled examples and prompt text for each language pair is included in our dataset release.

\subsection{Instructions for Computing Metrics}
\label{sec:metricdetails}
sentBLEU, chrF, chrF++, and TER scores were computed with sacreBLEU 2.4.2~\citep{post-2018-call} on python 3.11.8 with the following parameters:

chrF: -m chrf

chrF++: -m chrf --chrf-word-order 2

sentBLEU: -m bleu --sentence-level

TER: -m ter

For other metrics, we used the publicly released models on HuggingFace, running with the unbabel-comet package version 2.2.1 available on pip, on Python 3.10.14. We ran on an NVIDIA A100 GPU for all metrics except XCOMET-XXL and CometKiwi23-XXL, which required an NVIDIA A100 80GB GPU.

\section{Metrics Included in Each Ensemble}
\label{sec:ensemblemetrics}

This section presents the same information that is present in Table~\ref{tab:ensembles}, but in textual format. The following are the groups of metrics included in the single-step ensembles that we include in our study. For each of these metric groups the rankAvg, rankMed, rankMax, and rank75q ensembling techniques are used to generate an ensemble.

\begin{enumerate}[topsep=0pt,itemsep=-1ex,partopsep=1ex,parsep=1ex]
\item all: All metrics, both reference-based and QE (MetricX, MetricX-QE, XCOMET-XXL, XCOMET-XL, CometKiwi23-XXL, CometKiwi23-XL, CometKiwi22, COMET22, BLEURT, YiSi, chrF, chrF++, sentBLEU, TER, AfriCOMET and AfriCOMET-QE for African languages, IndicCOMET for Indic languages)
\item qe: All QE metrics (MetricX-QE, CometKiwi23-XXL, CometKiwi23-XL, CometKiwi22, and AfriCOMET-QE for African languages)
\item top: MetricX, MetricX-QE, XCOMET-XXL, XCOMET-XL, CometKiwi23-XXL, CometKiwi23-XL
\item topQe: MetricX-QE, CometKiwi23-XXL, CometKiwi23-XL
\item mxmxqe: MetricX, MetricX-QE
\item noLex: All non-lexical metrics (MetricX, MetricX-QE, XCOMET-XXL, XCOMET-XL, CometKiwi23-XXL, CometKiwi23-XL, CometKiwi22, COMET22, BLEURT, YiSi, AfriCOMET and AfriCOMET-QE for African languages, IndicCOMET for Indic languages)
\item noNC: All metrics that permit commercial use (MetricX, MetricX-QE, CometKiwi22, COMET22, BLEURT, YiSi, chrF, chrF++, sentBLEU, TER, AfriCOMET and AfriCOMET-QE for African languages, IndicCOMET for Indic languages)
\item noNCnoLex: All non-lexical metrics that permit commercial use (MetricX, MetricX-QE, COMET22, BLEURT, YiSi, AfriCOMET and AfriCOMET-QE for African languages, IndicCOMET for Indic languages)
\item noNCQe: All QE metrics that permit commercial use (MetricX-QE, and AfriCOMET-QE for African languages)
\end{enumerate}

In addition, we also investigate QE filtering followed by MBR decoding (here we define QE filtering as selecting the top N candidates according to a QE metric, where N can be either 4, 8, 16, 32, 64). We include the following ensembles of this form:

\begin{enumerate}[topsep=0pt,itemsep=-1ex,partopsep=1ex,parsep=1ex]
\item allQE(N)allMBR: Use QE filtering with an ensemble of all QE metrics (MetricX-QE, CometKiwi23-XXL, CometKiwi23-XL, CometKiwi22, AfriCOMET-QE for African languages), then perform MBR decoding on the N resulting candidates with all reference-based metrics (MetricX, XCOMET-XXL, XCOMET-XL, COMET22, BLEURT, YiSi, chrF, chrF++, sentBLEU, TER, AfriCOMET for African languages, IndicCOMET for Indic languages).
\item allQE(N)nolexMBR: Use QE filtering with an ensemble of all QE metrics (MetricX-QE, CometKiwi23-XXL, CometKiwi23-XL, CometKiwi22, AfriCOMET-QE for African languages), then perform MBR decoding on the N resulting candidates with all non-lexical reference-based metrics (MetricX, XCOMET-XXL, XCOMET-XL, COMET22, BLEURT, YiSi, AfriCOMET for African languages, IndicCOMET for Indic languages).
\item topQE(N)topMBR: Use QE filtering with an ensemble of top-performing QE metrics (MetricX QE, CometKiwi23-XXL, CometKiwi23-XL), then perform MBR decoding on the N resulting candidates with an ensemble of top-performing reference-based metrics (MetricX, XCOMET-XXL, XCOMET-XL).
\item noncQE(N)noncMBR: Use QE filtering with an ensemble of QE metrics that permit commercial use (MetricX-QE, AfriCOMET-QE for African languages), then perform MBR decoding with an ensemble of reference-based metrics that permit commercial use (MetricX, COMET22, BLEURT, YiSi, chrF, chrF++, sentBLEU, TER, AfriCOMET for African languages, IndicCOMET for Indic languages).
\item noncQE(N)noncnolexMBR: Use QE filtering with an ensemble of QE metrics that permit commercial use (MetricX-QE, AfriCOMET-QE for African languages), then perform MBR decoding with an ensemble of non-lexical reference-based metrics that permit commercial use (MetricX, COMET22, BLEURT, YiSi, AfriCOMET for African languages, IndicCOMET for Indic languages).
\item mxQE(N)xcMBR: Use QE filtering with MetricX-QE, then perform MBR decoding with XCOMET-XXL
\item ckQE(N)xcMBR: Use QE filtering with CometKiwi23-XXL, then perform MBR decoding with XCOMET-XXL
\item mxQE(N)mxMBR: Use QE filtering with MetricX-QE, then perform MBR decoding with MetricX
\item ckQE(N)mxMBR: Use QE filtering with CometKiwi23-XXL, then perform MBR decoding with MetricX
\end{enumerate}

\section{Pseudocode for Ensembles}
\label{sec:ensemblecode}

rankAvg ensembling strategy:

\begin{lstlisting}
def rankAvg(
  sample_list: List[str], metric_list: List[str]
):
  sample_ranks = get_ranks_for_samples_by_ensemble(sample_list, metric_list)
  score_list = [np.mean(x) for x in sample_ranks]
  return select_samples_by_score(sample_list, score_list)
\end{lstlisting}

rankMed ensembling strategy:

\begin{lstlisting}
def rankMed(
  sample_list: List[str], metric_list: List[str]
):
  sample_ranks = get_ranks_for_samples_by_ensemble(sample_list, metric_list)
  score_list = [np.median(x) for x in sample_ranks]
  return select_samples_by_score(sample_list, score_list)
\end{lstlisting}

rankMax ensembling strategy:

\begin{lstlisting}
def rankMax(
  sample_list: List[str], metric_list: List[str]
):
  sample_ranks = get_ranks_for_samples_by_ensemble(sample_list, metric_list)
  score_list = [np.max(x) for x in sample_ranks]
  return select_samples_by_score(sample_list, score_list)
\end{lstlisting}

rank75q ensembling strategy:

\begin{lstlisting}
def rank75q(
  sample_list: List[str], metric_list: List[str]
):
  sample_ranks = get_ranks_for_samples_by_ensemble(sample_list, metric_list)
  score_list = [np.quantile(x, q=[0.75])[0] for x in sample_ranks]
  return select_samples_by_score(sample_list, score_list)
\end{lstlisting}

Here are helper functions that were used:

\begin{lstlisting}
def get_ranks_for_samples_by_ensemble(
  sample_list: List[str], metric_list: List[str]
):
  output = [[None for y in metric_list] for x in sample_list]
  for metric_idx, metric in enumerate(metric_list):
    sample_to_rank = rank_samples_by_metric(sample_list, metric)
    for sample_idx, sample in enumerate(sample_list):
      output[sample_idx][metric_idx] = sample_to_rank[sample]
  return output

def select_samples_by_score(
  sample_list: List[str],
  score_list: List[float]
):
  sample_with_score = zip(sample_list, score_list)
  top_candidate, top_score = min(sample_with_score, key=lambda x: x[1])
  return top_candidate
\end{lstlisting}

\section{Correlation Between Human Evaluation MQM Scores and Metrics}

\label{sec:correlation}

\begin{table}[!hb]
\begin{small}
\setlength\tabcolsep{2pt}
\begin{tabular}{lrrrrrr}
& zh-en & en-de & en-ha & en-sw & en-hi & en-ml \\
XCOMET-XXL & {\cellcolor[HTML]{4BA54A}} \color[HTML]{F1F1F1} 0.391 & {\cellcolor[HTML]{BEDDBE}} \color[HTML]{000000} 0.084 & {\cellcolor[HTML]{41A041}} \color[HTML]{F1F1F1} 0.146 & {\cellcolor[HTML]{4CA54C}} \color[HTML]{F1F1F1} 0.139 & {\cellcolor[HTML]{8DC58D}} \color[HTML]{000000} 0.111 & {\cellcolor[HTML]{359A35}} \color[HTML]{F1F1F1} 0.202 \\
\hline
\scriptsize XCOMET-XXL:mbr & {\cellcolor[HTML]{4CA54C}} \color[HTML]{F1F1F1} 0.389 & {\cellcolor[HTML]{CAE3CA}} \color[HTML]{000000} 0.076 & {\cellcolor[HTML]{088408}} \color[HTML]{F1F1F1} 0.178 & {\cellcolor[HTML]{3F9F3F}} \color[HTML]{F1F1F1} 0.145 & {\cellcolor[HTML]{42A042}} \color[HTML]{F1F1F1} 0.141 & {\cellcolor[HTML]{3C9D3C}} \color[HTML]{F1F1F1} 0.198 \\
\hline
XCOMET-XL & {\cellcolor[HTML]{038103}} \color[HTML]{F1F1F1} 0.543 & {\cellcolor[HTML]{7EBE7E}} \color[HTML]{000000} 0.126 & {\cellcolor[HTML]{329832}} \color[HTML]{F1F1F1} 0.154 & {\cellcolor[HTML]{219021}} \color[HTML]{F1F1F1} 0.160 & {\cellcolor[HTML]{40A040}} \color[HTML]{F1F1F1} 0.141 & {\cellcolor[HTML]{2C962C}} \color[HTML]{F1F1F1} 0.208 \\
\hline
\scriptsize XCOMET-XL:mbr & {\cellcolor[HTML]{008000}} \color[HTML]{F1F1F1} 0.550 & {\cellcolor[HTML]{82C082}} \color[HTML]{000000} 0.124 & {\cellcolor[HTML]{178B17}} \color[HTML]{F1F1F1} 0.170 & {\cellcolor[HTML]{048204}} \color[HTML]{F1F1F1} 0.174 & {\cellcolor[HTML]{1B8D1B}} \color[HTML]{F1F1F1} 0.156 & {\cellcolor[HTML]{42A042}} \color[HTML]{F1F1F1} 0.194 \\
\hline
MetricX & {\cellcolor[HTML]{4BA54B}} \color[HTML]{F1F1F1} 0.391 & {\cellcolor[HTML]{9ECE9E}} \color[HTML]{000000} 0.105 & {\cellcolor[HTML]{BADBBA}} \color[HTML]{000000} 0.077 & {\cellcolor[HTML]{3E9E3E}} \color[HTML]{F1F1F1} 0.146 & {\cellcolor[HTML]{A9D3A9}} \color[HTML]{000000} 0.100 & {\cellcolor[HTML]{1E8F1E}} \color[HTML]{F1F1F1} 0.216 \\
\hline
MetricX:mbr & {\cellcolor[HTML]{389B38}} \color[HTML]{F1F1F1} 0.431 & {\cellcolor[HTML]{7DBD7D}} \color[HTML]{000000} 0.127 & {\cellcolor[HTML]{118911}} \color[HTML]{F1F1F1} 0.173 & {\cellcolor[HTML]{359A35}} \color[HTML]{F1F1F1} 0.150 & {\cellcolor[HTML]{239123}} \color[HTML]{F1F1F1} 0.153 & {\cellcolor[HTML]{389B38}} \color[HTML]{F1F1F1} 0.200 \\
\hline
MetricX-QE & {\cellcolor[HTML]{1E8F1E}} \color[HTML]{F1F1F1} 0.485 & {\cellcolor[HTML]{87C287}} \color[HTML]{000000} 0.120 & {\cellcolor[HTML]{77BA77}} \color[HTML]{F1F1F1} 0.115 & {\cellcolor[HTML]{5BAD5B}} \color[HTML]{F1F1F1} 0.132 & {\cellcolor[HTML]{E9F2E9}} \color[HTML]{000000} 0.074 & {\cellcolor[HTML]{6AB46A}} \color[HTML]{F1F1F1} 0.170 \\
\hline
\scriptsize CometKiwi23-XXL & {\cellcolor[HTML]{92C892}} \color[HTML]{000000} 0.241 & {\cellcolor[HTML]{B9DBB9}} \color[HTML]{000000} 0.088 & {\cellcolor[HTML]{76BA76}} \color[HTML]{F1F1F1} 0.116 & {\cellcolor[HTML]{72B872}} \color[HTML]{F1F1F1} 0.121 & {\cellcolor[HTML]{62B062}} \color[HTML]{F1F1F1} 0.128 & {\cellcolor[HTML]{2B952B}} \color[HTML]{F1F1F1} 0.208 \\
\hline
\scriptsize CometKiwi23-XL & {\cellcolor[HTML]{7EBE7E}} \color[HTML]{000000} 0.284 & {\cellcolor[HTML]{B2D7B2}} \color[HTML]{000000} 0.092 & {\cellcolor[HTML]{96C996}} \color[HTML]{000000} 0.098 & {\cellcolor[HTML]{78BB78}} \color[HTML]{F1F1F1} 0.118 & {\cellcolor[HTML]{6CB56C}} \color[HTML]{F1F1F1} 0.124 & {\cellcolor[HTML]{359A35}} \color[HTML]{F1F1F1} 0.202 \\
\hline
CometKiwi22 & {\cellcolor[HTML]{82C082}} \color[HTML]{000000} 0.277 & {\cellcolor[HTML]{5CAD5C}} \color[HTML]{F1F1F1} 0.148 & {\cellcolor[HTML]{EBF3EB}} \color[HTML]{000000} 0.050 & {\cellcolor[HTML]{299429}} \color[HTML]{F1F1F1} 0.156 & {\cellcolor[HTML]{81BF81}} \color[HTML]{000000} 0.116 & {\cellcolor[HTML]{008000}} \color[HTML]{F1F1F1} 0.235 \\
\hline
COMET22 & {\cellcolor[HTML]{78BB78}} \color[HTML]{F1F1F1} 0.298 & {\cellcolor[HTML]{3B9D3B}} \color[HTML]{F1F1F1} 0.170 & {\cellcolor[HTML]{DDECDD}} \color[HTML]{000000} 0.058 & {\cellcolor[HTML]{3F9F3F}} \color[HTML]{F1F1F1} 0.146 & {\cellcolor[HTML]{ABD4AB}} \color[HTML]{000000} 0.099 & {\cellcolor[HTML]{41A041}} \color[HTML]{F1F1F1} 0.195 \\
\hline
COMET22:mbr & {\cellcolor[HTML]{70B770}} \color[HTML]{F1F1F1} 0.312 & {\cellcolor[HTML]{008000}} \color[HTML]{F1F1F1} 0.209 & {\cellcolor[HTML]{C9E3C9}} \color[HTML]{000000} 0.069 & {\cellcolor[HTML]{389B38}} \color[HTML]{F1F1F1} 0.149 & {\cellcolor[HTML]{7BBC7B}} \color[HTML]{000000} 0.118 & {\cellcolor[HTML]{49A449}} \color[HTML]{F1F1F1} 0.190 \\
\hline
BLEURT & {\cellcolor[HTML]{72B872}} \color[HTML]{F1F1F1} 0.308 & {\cellcolor[HTML]{56AA56}} \color[HTML]{F1F1F1} 0.152 & {\cellcolor[HTML]{56AA56}} \color[HTML]{F1F1F1} 0.134 & {\cellcolor[HTML]{43A143}} \color[HTML]{F1F1F1} 0.143 & {\cellcolor[HTML]{82C082}} \color[HTML]{000000} 0.115 & {\cellcolor[HTML]{309730}} \color[HTML]{F1F1F1} 0.205 \\
\hline
BLEURT:mbr & {\cellcolor[HTML]{6CB56C}} \color[HTML]{F1F1F1} 0.322 & {\cellcolor[HTML]{3B9D3B}} \color[HTML]{F1F1F1} 0.170 & {\cellcolor[HTML]{47A347}} \color[HTML]{F1F1F1} 0.143 & {\cellcolor[HTML]{369B36}} \color[HTML]{F1F1F1} 0.150 & {\cellcolor[HTML]{2C962C}} \color[HTML]{F1F1F1} 0.149 & {\cellcolor[HTML]{48A348}} \color[HTML]{F1F1F1} 0.191 \\
\hline
YiSi & {\cellcolor[HTML]{A0CEA0}} \color[HTML]{000000} 0.211 & {\cellcolor[HTML]{B9DBB9}} \color[HTML]{000000} 0.088 & {\cellcolor[HTML]{89C389}} \color[HTML]{000000} 0.105 & {\cellcolor[HTML]{9BCC9B}} \color[HTML]{000000} 0.100 & {\cellcolor[HTML]{BBDCBB}} \color[HTML]{000000} 0.092 & {\cellcolor[HTML]{4EA64E}} \color[HTML]{F1F1F1} 0.187 \\
\hline
YiSi:mbr & {\cellcolor[HTML]{9FCE9F}} \color[HTML]{000000} 0.214 & {\cellcolor[HTML]{82C082}} \color[HTML]{000000} 0.124 & {\cellcolor[HTML]{4FA74F}} \color[HTML]{F1F1F1} 0.138 & {\cellcolor[HTML]{92C892}} \color[HTML]{000000} 0.105 & {\cellcolor[HTML]{A7D2A7}} \color[HTML]{000000} 0.100 & {\cellcolor[HTML]{369B36}} \color[HTML]{F1F1F1} 0.202 \\
\hline
chrF & {\cellcolor[HTML]{EBF3EB}} \color[HTML]{000000} 0.054 & {\cellcolor[HTML]{EBF3EB}} \color[HTML]{000000} 0.055 & {\cellcolor[HTML]{87C287}} \color[HTML]{000000} 0.106 & {\cellcolor[HTML]{90C790}} \color[HTML]{000000} 0.106 & {\cellcolor[HTML]{CAE3CA}} \color[HTML]{000000} 0.086 & {\cellcolor[HTML]{73B873}} \color[HTML]{F1F1F1} 0.164 \\
\hline
chrF:mbr & {\cellcolor[HTML]{DDECDD}} \color[HTML]{000000} 0.083 & {\cellcolor[HTML]{D5E9D5}} \color[HTML]{000000} 0.069 & {\cellcolor[HTML]{7FBE7F}} \color[HTML]{000000} 0.111 & {\cellcolor[HTML]{7DBD7D}} \color[HTML]{000000} 0.115 & {\cellcolor[HTML]{D0E6D0}} \color[HTML]{000000} 0.084 & {\cellcolor[HTML]{4BA54A}} \color[HTML]{F1F1F1} 0.189 \\
\hline
chrF++ & {\cellcolor[HTML]{E7F1E7}} \color[HTML]{000000} 0.062 & {\cellcolor[HTML]{E7F1E7}} \color[HTML]{000000} 0.058 & {\cellcolor[HTML]{82C082}} \color[HTML]{000000} 0.109 & {\cellcolor[HTML]{8DC58D}} \color[HTML]{000000} 0.108 & {\cellcolor[HTML]{C9E3C9}} \color[HTML]{000000} 0.087 & {\cellcolor[HTML]{7FBE7F}} \color[HTML]{000000} 0.157 \\
\hline
chrF++:mbr & {\cellcolor[HTML]{D9EAD9}} \color[HTML]{000000} 0.091 & {\cellcolor[HTML]{D5E9D5}} \color[HTML]{000000} 0.069 & {\cellcolor[HTML]{81BF81}} \color[HTML]{000000} 0.110 & {\cellcolor[HTML]{80BF80}} \color[HTML]{000000} 0.113 & {\cellcolor[HTML]{C5E0C5}} \color[HTML]{000000} 0.088 & {\cellcolor[HTML]{55A955}} \color[HTML]{F1F1F1} 0.183 \\
\hline
sentBLEU & {\cellcolor[HTML]{C9E2C8}} \color[HTML]{000000} 0.128 & {\cellcolor[HTML]{D1E6D1}} \color[HTML]{000000} 0.072 & {\cellcolor[HTML]{9BCC9B}} \color[HTML]{000000} 0.095 & {\cellcolor[HTML]{A8D2A8}} \color[HTML]{000000} 0.094 & {\cellcolor[HTML]{EBF3EB}} \color[HTML]{000000} 0.073 & {\cellcolor[HTML]{EBF3EB}} \color[HTML]{000000} 0.091 \\
\hline
sentBLEU:mbr & {\cellcolor[HTML]{B9DBB9}} \color[HTML]{000000} 0.160 & {\cellcolor[HTML]{D2E7D2}} \color[HTML]{000000} 0.072 & {\cellcolor[HTML]{96C996}} \color[HTML]{000000} 0.098 & {\cellcolor[HTML]{86C286}} \color[HTML]{000000} 0.111 & {\cellcolor[HTML]{C6E1C6}} \color[HTML]{000000} 0.088 & {\cellcolor[HTML]{C7E1C7}} \color[HTML]{000000} 0.113 \\
\hline
TER & {\cellcolor[HTML]{D4E8D4}} \color[HTML]{000000} 0.101 & {\cellcolor[HTML]{D2E7D2}} \color[HTML]{000000} 0.071 & {\cellcolor[HTML]{8AC48A}} \color[HTML]{000000} 0.104 & {\cellcolor[HTML]{EBF3EB}} \color[HTML]{000000} 0.061 & {\cellcolor[HTML]{9CCD9C}} \color[HTML]{000000} 0.105 & {\cellcolor[HTML]{BFDEBF}} \color[HTML]{000000} 0.118 \\
\hline
TER:mbr & {\cellcolor[HTML]{D7E9D7}} \color[HTML]{000000} 0.096 & {\cellcolor[HTML]{BDDCBD}} \color[HTML]{000000} 0.085 & {\cellcolor[HTML]{89C389}} \color[HTML]{000000} 0.105 & {\cellcolor[HTML]{DFEDDF}} \color[HTML]{000000} 0.067 & {\cellcolor[HTML]{97CA97}} \color[HTML]{000000} 0.107 & {\cellcolor[HTML]{8CC58C}} \color[HTML]{000000} 0.149 \\
\hline
\vtop{\hbox{\strut MetricX}\hbox{\strut +MetricX-QE}}  & {\cellcolor[HTML]{289428}} \color[HTML]{F1F1F1} 0.463 & {\cellcolor[HTML]{81BF81}} \color[HTML]{000000} 0.124 & {\cellcolor[HTML]{90C790}} \color[HTML]{000000} 0.101 & {\cellcolor[HTML]{319831}} \color[HTML]{F1F1F1} 0.152 & {\cellcolor[HTML]{A5D1A5}} \color[HTML]{000000} 0.101 & {\cellcolor[HTML]{098509}} \color[HTML]{F1F1F1} 0.229 \\
\hline
\vtop{\hbox{\strut MetricX}\hbox{\strut +MetricX-QE}} & {\cellcolor[HTML]{1F8F1F}} \color[HTML]{F1F1F1} 0.483 & {\cellcolor[HTML]{78BB78}} \color[HTML]{000000} 0.130 & {\cellcolor[HTML]{289428}} \color[HTML]{F1F1F1} 0.160 & {\cellcolor[HTML]{349A34}} \color[HTML]{F1F1F1} 0.151 & {\cellcolor[HTML]{5BAD5B}} \color[HTML]{F1F1F1} 0.131 & {\cellcolor[HTML]{2A952A}} \color[HTML]{F1F1F1} 0.209 \\
\hline
\vtop{\hbox{\strut XCOMET-XXL}\hbox{\strut +XCOMET-XL}} & {\cellcolor[HTML]{088408}} \color[HTML]{F1F1F1} 0.532 & {\cellcolor[HTML]{97CA97}} \color[HTML]{000000} 0.110 & {\cellcolor[HTML]{2A952A}} \color[HTML]{F1F1F1} 0.159 & {\cellcolor[HTML]{1E8F1E}} \color[HTML]{F1F1F1} 0.161 & {\cellcolor[HTML]{399C39}} \color[HTML]{F1F1F1} 0.144 & {\cellcolor[HTML]{0C860C}} \color[HTML]{F1F1F1} 0.228 \\
\hline
\vtop{\hbox{\strut \scriptsize XCOMET-XXL:mbr}\hbox{\strut \scriptsize +XCOMET-XL:mbr}} & {\cellcolor[HTML]{068306}} \color[HTML]{F1F1F1} 0.537 & {\cellcolor[HTML]{9FCE9F}} \color[HTML]{000000} 0.105 & {\cellcolor[HTML]{008000}} \color[HTML]{F1F1F1} 0.183 & {\cellcolor[HTML]{0A850A}} \color[HTML]{F1F1F1} 0.171 & {\cellcolor[HTML]{018001}} \color[HTML]{F1F1F1} 0.166 & {\cellcolor[HTML]{209020}} \color[HTML]{F1F1F1} 0.215 \\
\hline
\vtop{\hbox{\strut XCOMET-XXL}\hbox{\strut +XCOMET-XL}\hbox{\strut +COMET22}} & {\cellcolor[HTML]{0E870E}} \color[HTML]{F1F1F1} 0.521 & {\cellcolor[HTML]{6FB76F}} \color[HTML]{F1F1F1} 0.136 & {\cellcolor[HTML]{399C39}} \color[HTML]{F1F1F1} 0.150 & {\cellcolor[HTML]{0E870E}} \color[HTML]{F1F1F1} 0.169 & {\cellcolor[HTML]{389B38}} \color[HTML]{F1F1F1} 0.144 & {\cellcolor[HTML]{008000}} \color[HTML]{F1F1F1} 0.235 \\
\hline
\vtop{\hbox{\strut \scriptsize XCOMET-XXL:mbr}\hbox{\strut \scriptsize +XCOMET-XL:mbr}\hbox{\strut +COMET22:mbr}} & {\cellcolor[HTML]{098509}} \color[HTML]{F1F1F1} 0.529 & {\cellcolor[HTML]{6EB66E}} \color[HTML]{F1F1F1} 0.136 & {\cellcolor[HTML]{118911}} \color[HTML]{F1F1F1} 0.173 & {\cellcolor[HTML]{008000}} \color[HTML]{F1F1F1} 0.176 & {\cellcolor[HTML]{008000}} \color[HTML]{F1F1F1} 0.167 & {\cellcolor[HTML]{128912}} \color[HTML]{F1F1F1} 0.223 \\
\end{tabular}
\end{small}
\caption{Pearson correlation between MQM evaluation scores and automated evaluation scores. For reference-based metrics, rows with ``:mbr'' indicate pseudoreference-based evaluation. Bottom rows are ensembles that take the average between the listed metrics. Higher scores indicate better agreement with human raters. See Table~\ref{tab:metric_correlation_kendalltau} for Kendall-Tau correlation.}
\label{tab:metric_correlation_pearson}
\end{table}


\onecolumn

\clearpage

\section{Results on Dev Datasets (WMT2022 and FLORES200 dev)}

\label{sec:devresults}

\begin{table}[h!]
\makebox[\textwidth]{
\setlength\tabcolsep{2 pt}
\begin{small}

\begin{tabular}{lcccccccccccccc}
\textbf{MBR/QE Method} \hfill \hfill \hfill \hfill \hfill \hfill \hfill \hfill \hfill \hfill \lrot{\textbf{Evaluated Metric}} & \lrot{MetricX} & \lrot{MetricX-QE} & \lrot{XCOMET-XXL} & \lrot{XCOMET-XL} & \lrot{CometKiwi23-XXL} & \lrot{CometKiwi23-XL} & \lrot{CometKiwi22} & \lrot{COMET22} & \lrot{BLEURT} & \lrot{YiSi} & \lrot{chrF} & \lrot{chrF++} & \lrot{sentBLEU} & \lrot{TER} \\
Greedy & {\cellcolor[HTML]{FDFFFD}} \color[HTML]{000000} 1.58 & {\cellcolor[HTML]{FDFFFD}} \color[HTML]{000000} 1.16 & {\cellcolor[HTML]{FDFFFD}} \color[HTML]{000000} 82.3 & {\cellcolor[HTML]{FDFFFD}} \color[HTML]{000000} 77.8 & {\cellcolor[HTML]{FDFFFD}} \color[HTML]{000000} 76.8 & {\cellcolor[HTML]{FDFFFD}} \color[HTML]{000000} 68.2 & {\cellcolor[HTML]{FDFFFD}} \color[HTML]{000000} 77.5 & {\cellcolor[HTML]{FDFFFD}} \color[HTML]{000000} 85.2 & {\cellcolor[HTML]{FDFFFD}} \color[HTML]{000000} 77.3 & {\cellcolor[HTML]{FDFFFD}} \color[HTML]{000000} 84.3 & {\cellcolor[HTML]{FDFFFD}} \color[HTML]{000000} 57.2 & {\cellcolor[HTML]{FDFFFD}} \color[HTML]{000000} 54.2 & {\cellcolor[HTML]{FDFFFD}} \color[HTML]{000000} 26.4 & {\cellcolor[HTML]{FDFFFD}} \color[HTML]{000000} 63.4 \\
\end{tabular}

\end{small}
}
\begin{small}

\caption{Reference-based and QE evaluation scores for greedy, MBR, and QE decoding using a single-step ensemble utility metric, averaged across all languages (test datasets). Higher scores are better, except MetricX, MetricX-QE, and TER, where lower is better. Green is better than greedy, red is worse.  Ensembles are defined in Table~\ref{tab:ensembles}. Significant differences from greedy (pairwise t-test) indicated by * for p<0.05, † for p<0.01, ‡ for p<0.001.}
\label{tab:singlestep_all_test}
\end{small}
\end{table}

\clearpage

\section{Results for Additional Ensembles}

\label{sec:extraensembles}

\subsection{Additional Single-Step Ensembles on Test Datasets}

\label{sec:extrasinglestep}

\begin{table}[h!]
\makebox[\textwidth]{
\setlength\tabcolsep{2 pt}
\begin{small}

\begin{tabular}{lcccccccccccccc}
\textbf{MBR/QE Method} \hfill \hfill \hfill \hfill \hfill \hfill \hfill \hfill \hfill \hfill \lrot{\textbf{Evaluated Metric}} & \lrot{MetricX} & \lrot{MetricX-QE} & \lrot{XCOMET-XXL} & \lrot{XCOMET-XL} & \lrot{CometKiwi23-XXL} & \lrot{CometKiwi23-XL} & \lrot{CometKiwi22} & \lrot{COMET22} & \lrot{BLEURT} & \lrot{YiSi} & \lrot{chrF} & \lrot{chrF++} & \lrot{sentBLEU} & \lrot{TER} \\
Greedy & {\cellcolor[HTML]{FDFFFD}} \color[HTML]{000000} 1.58 & {\cellcolor[HTML]{FDFFFD}} \color[HTML]{000000} 1.16 & {\cellcolor[HTML]{FDFFFD}} \color[HTML]{000000} 82.3 & {\cellcolor[HTML]{FDFFFD}} \color[HTML]{000000} 77.8 & {\cellcolor[HTML]{FDFFFD}} \color[HTML]{000000} 76.8 & {\cellcolor[HTML]{FDFFFD}} \color[HTML]{000000} 68.2 & {\cellcolor[HTML]{FDFFFD}} \color[HTML]{000000} 77.5 & {\cellcolor[HTML]{FDFFFD}} \color[HTML]{000000} 85.2 & {\cellcolor[HTML]{FDFFFD}} \color[HTML]{000000} 77.3 & {\cellcolor[HTML]{FDFFFD}} \color[HTML]{000000} 84.3 & {\cellcolor[HTML]{FDFFFD}} \color[HTML]{000000} 57.2 & {\cellcolor[HTML]{FDFFFD}} \color[HTML]{000000} 54.2 & {\cellcolor[HTML]{FDFFFD}} \color[HTML]{000000} 26.4 & {\cellcolor[HTML]{FDFFFD}} \color[HTML]{000000} 63.4 \\
\end{tabular}

\end{small}
}
\begin{small}

\caption{Reference-based and QE evaluation scores for greedy, MBR, and QE decoding using a single-step ensemble utility metric, averaged across all languages (test datasets). Higher scores are better, except MetricX, MetricX-QE, and TER, where lower is better. Green is better than greedy, red is worse.  Ensembles are defined in Table~\ref{tab:ensembles}. Significant differences from greedy (pairwise t-test) indicated by * for p<0.05, † for p<0.01, ‡ for p<0.001.}
\label{tab:singlestep_all_test}
\end{small}
\end{table}

\clearpage

\subsection{Additional Two-Step Ensembles on Test Datasets}

\label{sec:extratwostep}

\begin{table}[h!]
\makebox[\textwidth]{
\setlength\tabcolsep{2 pt}
\begin{small}

\begin{tabular}{lcccccccccccccc}
\textbf{MBR/QE Method} \hfill \hfill \hfill \hfill \hfill \hfill \hfill \hfill \hfill \hfill \lrot{\textbf{Evaluated Metric}} & \lrot{MetricX} & \lrot{MetricX-QE} & \lrot{XCOMET-XXL} & \lrot{XCOMET-XL} & \lrot{CometKiwi23-XXL} & \lrot{CometKiwi23-XL} & \lrot{CometKiwi22} & \lrot{COMET22} & \lrot{BLEURT} & \lrot{YiSi} & \lrot{chrF} & \lrot{chrF++} & \lrot{sentBLEU} & \lrot{TER} \\
Greedy & {\cellcolor[HTML]{FDFFFD}} \color[HTML]{000000} 1.58 & {\cellcolor[HTML]{FDFFFD}} \color[HTML]{000000} 1.16 & {\cellcolor[HTML]{FDFFFD}} \color[HTML]{000000} 82.3 & {\cellcolor[HTML]{FDFFFD}} \color[HTML]{000000} 77.8 & {\cellcolor[HTML]{FDFFFD}} \color[HTML]{000000} 76.8 & {\cellcolor[HTML]{FDFFFD}} \color[HTML]{000000} 68.2 & {\cellcolor[HTML]{FDFFFD}} \color[HTML]{000000} 77.5 & {\cellcolor[HTML]{FDFFFD}} \color[HTML]{000000} 85.2 & {\cellcolor[HTML]{FDFFFD}} \color[HTML]{000000} 77.3 & {\cellcolor[HTML]{FDFFFD}} \color[HTML]{000000} 84.3 & {\cellcolor[HTML]{FDFFFD}} \color[HTML]{000000} 57.2 & {\cellcolor[HTML]{FDFFFD}} \color[HTML]{000000} 54.2 & {\cellcolor[HTML]{FDFFFD}} \color[HTML]{000000} 26.4 & {\cellcolor[HTML]{FDFFFD}} \color[HTML]{000000} 63.4 \\
\end{tabular}

\end{small}
}
\begin{small}

\caption{Reference-based and QE evaluation scores for greedy, MBR, and QE decoding using a two-step ensemble (QE filtering followed by MBR) utility metric, averaged across all languages (test datasets). Higher scores are better, except MetricX, MetricX-QE, and TER, where lower is better. Green is better than greedy, red is worse.  Ensembles are defined in Table~\ref{tab:ensembles}. Significant differences from greedy (pairwise t-test) indicated by * for p<0.05, † for p<0.01, ‡ for p<0.001.}
\label{tab:multistep_all_test}
\end{small}
\end{table}

\clearpage

\section{Breakdown of Results on Individual Language Pairs}
\label{sec:perlangresults}

\subsection{Results for English-Swahili (en-sw) on FLORES200 test dataset}
\begin{table}[h!]

\makebox[\textwidth]{
\setlength\tabcolsep{2 pt}
\begin{small}

\begin{tabular}{lcccccccccccccccc}
\textbf{MBR/QE Method} \hfill \hfill \hfill \hfill \hfill \hfill \hfill \hfill \hfill \hfill \lrot{\textbf{Evaluated Metric}} & \lrot{MetricX} & \lrot{MetricX-QE} & \lrot{XCOMET-XXL} & \lrot{XCOMET-XL} & \lrot{CometKiwi23-XXL} & \lrot{CometKiwi23-XL} & \lrot{CometKiwi22} & \lrot{COMET22} & \lrot{AfriCOMET} & \lrot{AfriCOMET-QE} & \lrot{BLEURT} & \lrot{YiSi} & \lrot{chrF} & \lrot{chrF++} & \lrot{sentBLEU} & \lrot{TER} \\
Greedy & {\cellcolor[HTML]{FDFFFD}} \color[HTML]{000000} 1.70 & {\cellcolor[HTML]{FDFFFD}} \color[HTML]{000000} 1.28 & {\cellcolor[HTML]{FDFFFD}} \color[HTML]{000000} 83.7 & {\cellcolor[HTML]{FDFFFD}} \color[HTML]{000000} 85.0 & {\cellcolor[HTML]{FDFFFD}} \color[HTML]{000000} 84.6 & {\cellcolor[HTML]{FDFFFD}} \color[HTML]{000000} 73.2 & {\cellcolor[HTML]{FDFFFD}} \color[HTML]{000000} 83.8 & {\cellcolor[HTML]{FDFFFD}} \color[HTML]{000000} 86.0 & {\cellcolor[HTML]{FDFFFD}} \color[HTML]{000000} 85.7 & {\cellcolor[HTML]{FDFFFD}} \color[HTML]{000000} 76.9 & {\cellcolor[HTML]{FDFFFD}} \color[HTML]{000000} 77.5 & {\cellcolor[HTML]{FDFFFD}} \color[HTML]{000000} 86.3 & {\cellcolor[HTML]{FDFFFD}} \color[HTML]{000000} 65.0 & {\cellcolor[HTML]{FDFFFD}} \color[HTML]{000000} 62.6 & {\cellcolor[HTML]{FDFFFD}} \color[HTML]{000000} 34.9 & {\cellcolor[HTML]{FDFFFD}} \color[HTML]{000000} 51.7 \\
\end{tabular}

\end{small}
}
\begin{small}

\caption{Reference-based and QE evaluation scores for greedy and MBR/QE decoding (1\textsuperscript{st} block), and ensembles (2\textsuperscript{nd} block), on en-sw (FLORES200 test dataset). Higher scores are better, except MetricX, MetricX-QE, and TER, where lower is better. Green is better than greedy, red is worse.  Ensembles are defined in Table~\ref{tab:ensembles}. Significant differences from greedy (pairwise t-test) indicated by * for p<0.05, † for p<0.01, ‡ for p<0.001. The green diagonal in the 1\textsuperscript{st} block shows metrics prefer outputs from MBR/QE decoding using the same utility metric.}
\label{tab:subset_en_sw_test}
\end{small}
    
\end{table}
\clearpage

\subsection{Results for English-Hausa (en-ha) on FLORES200 test dataset}
\begin{table}[h!]

\makebox[\textwidth]{
\setlength\tabcolsep{2 pt}
\begin{small}

\begin{tabular}{lcccccccccccccccc}
\textbf{MBR/QE Method} \hfill \hfill \hfill \hfill \hfill \hfill \hfill \hfill \hfill \hfill \lrot{\textbf{Evaluated Metric}} & \lrot{MetricX} & \lrot{MetricX-QE} & \lrot{XCOMET-XXL} & \lrot{XCOMET-XL} & \lrot{CometKiwi23-XXL} & \lrot{CometKiwi23-XL} & \lrot{CometKiwi22} & \lrot{COMET22} & \lrot{AfriCOMET} & \lrot{AfriCOMET-QE} & \lrot{BLEURT} & \lrot{YiSi} & \lrot{chrF} & \lrot{chrF++} & \lrot{sentBLEU} & \lrot{TER} \\
Greedy & {\cellcolor[HTML]{FDFFFD}} \color[HTML]{000000} 2.31 & {\cellcolor[HTML]{FDFFFD}} \color[HTML]{000000} 1.46 & {\cellcolor[HTML]{FDFFFD}} \color[HTML]{000000} 74.7 & {\cellcolor[HTML]{FDFFFD}} \color[HTML]{000000} 70.3 & {\cellcolor[HTML]{FDFFFD}} \color[HTML]{000000} 79.9 & {\cellcolor[HTML]{FDFFFD}} \color[HTML]{000000} 66.2 & {\cellcolor[HTML]{FDFFFD}} \color[HTML]{000000} 60.1 & {\cellcolor[HTML]{FDFFFD}} \color[HTML]{000000} 81.5 & {\cellcolor[HTML]{FDFFFD}} \color[HTML]{000000} 79.3 & {\cellcolor[HTML]{FDFFFD}} \color[HTML]{000000} 72.3 & {\cellcolor[HTML]{FDFFFD}} \color[HTML]{000000} 84.1 & {\cellcolor[HTML]{FDFFFD}} \color[HTML]{000000} 79.5 & {\cellcolor[HTML]{FDFFFD}} \color[HTML]{000000} 52.5 & {\cellcolor[HTML]{FDFFFD}} \color[HTML]{000000} 49.8 & {\cellcolor[HTML]{FDFFFD}} \color[HTML]{000000} 21.3 & {\cellcolor[HTML]{FDFFFD}} \color[HTML]{000000} 65.9 \\
\end{tabular}

\end{small}
}
\begin{small}

\caption{Reference-based and QE evaluation scores for greedy and MBR/QE decoding (1\textsuperscript{st} block), and ensembles (2\textsuperscript{nd} block), on en-ha (FLORES200 test dataset). Higher scores are better, except MetricX, MetricX-QE, and TER, where lower is better. Green is better than greedy, red is worse.  Ensembles are defined in Table~\ref{tab:ensembles}. Significant differences from greedy (pairwise t-test) indicated by * for p<0.05, † for p<0.01, ‡ for p<0.001. The green diagonal in the 1\textsuperscript{st} block shows metrics prefer outputs from MBR/QE decoding using the same utility metric.}
\label{tab:subset_en_ha_test}
\end{small}
    
\end{table}
\clearpage

\subsection{Results for English-Igbo (en-ig) on FLORES200 test dataset}
\begin{table}[h!]

\makebox[\textwidth]{
\setlength\tabcolsep{2 pt}
\begin{small}

\begin{tabular}{lcccccccccccccccc}
\textbf{MBR/QE Method} \hfill \hfill \hfill \hfill \hfill \hfill \hfill \hfill \hfill \hfill \lrot{\textbf{Evaluated Metric}} & \lrot{MetricX} & \lrot{MetricX-QE} & \lrot{XCOMET-XXL} & \lrot{XCOMET-XL} & \lrot{CometKiwi23-XXL} & \lrot{CometKiwi23-XL} & \lrot{CometKiwi22} & \lrot{COMET22} & \lrot{AfriCOMET} & \lrot{AfriCOMET-QE} & \lrot{BLEURT} & \lrot{YiSi} & \lrot{chrF} & \lrot{chrF++} & \lrot{sentBLEU} & \lrot{TER} \\
Greedy & {\cellcolor[HTML]{FDFFFD}} \color[HTML]{000000} 3.73 & {\cellcolor[HTML]{FDFFFD}} \color[HTML]{000000} 2.67 & {\cellcolor[HTML]{FDFFFD}} \color[HTML]{000000} 27.3 & {\cellcolor[HTML]{FDFFFD}} \color[HTML]{000000} 19.8 & {\cellcolor[HTML]{FDFFFD}} \color[HTML]{000000} 16.9 & {\cellcolor[HTML]{FDFFFD}} \color[HTML]{000000} 16.8 & {\cellcolor[HTML]{FDFFFD}} \color[HTML]{000000} 30.2 & {\cellcolor[HTML]{FDFFFD}} \color[HTML]{000000} 72.1 & {\cellcolor[HTML]{FDFFFD}} \color[HTML]{000000} 68.9 & {\cellcolor[HTML]{FDFFFD}} \color[HTML]{000000} 66.4 & {\cellcolor[HTML]{FDFFFD}} \color[HTML]{000000} 33.5 & {\cellcolor[HTML]{FDFFFD}} \color[HTML]{000000} 78.3 & {\cellcolor[HTML]{FDFFFD}} \color[HTML]{000000} 42.7 & {\cellcolor[HTML]{FDFFFD}} \color[HTML]{000000} 40.7 & {\cellcolor[HTML]{FDFFFD}} \color[HTML]{000000} 15.2 & {\cellcolor[HTML]{FDFFFD}} \color[HTML]{000000} 74.2 \\
\end{tabular}

\end{small}
}
\begin{small}

\caption{Reference-based and QE evaluation scores for greedy and MBR/QE decoding (1\textsuperscript{st} block), and ensembles (2\textsuperscript{nd} block), on en-ig (FLORES200 test dataset). Higher scores are better, except MetricX, MetricX-QE, and TER, where lower is better. Green is better than greedy, red is worse.  Ensembles are defined in Table~\ref{tab:ensembles}. Significant differences from greedy (pairwise t-test) indicated by * for p<0.05, † for p<0.01, ‡ for p<0.001. The green diagonal in the 1\textsuperscript{st} block shows metrics prefer outputs from MBR/QE decoding using the same utility metric.}
\label{tab:subset_en_ig_test}
\end{small}
    
\end{table}
\clearpage

\subsection{Results for English-Somali (en-so) on FLORES200 test dataset}
\begin{table}[h!]

\makebox[\textwidth]{
\setlength\tabcolsep{2 pt}
\begin{small}

\begin{tabular}{lcccccccccccccccc}
\textbf{MBR/QE Method} \hfill \hfill \hfill \hfill \hfill \hfill \hfill \hfill \hfill \hfill \lrot{\textbf{Evaluated Metric}} & \lrot{MetricX} & \lrot{MetricX-QE} & \lrot{XCOMET-XXL} & \lrot{XCOMET-XL} & \lrot{CometKiwi23-XXL} & \lrot{CometKiwi23-XL} & \lrot{CometKiwi22} & \lrot{COMET22} & \lrot{AfriCOMET} & \lrot{AfriCOMET-QE} & \lrot{BLEURT} & \lrot{YiSi} & \lrot{chrF} & \lrot{chrF++} & \lrot{sentBLEU} & \lrot{TER} \\
Greedy & {\cellcolor[HTML]{FDFFFD}} \color[HTML]{000000} 2.66 & {\cellcolor[HTML]{FDFFFD}} \color[HTML]{000000} 1.89 & {\cellcolor[HTML]{FDFFFD}} \color[HTML]{000000} 67.9 & {\cellcolor[HTML]{FDFFFD}} \color[HTML]{000000} 66.0 & {\cellcolor[HTML]{FDFFFD}} \color[HTML]{000000} 78.3 & {\cellcolor[HTML]{FDFFFD}} \color[HTML]{000000} 65.8 & {\cellcolor[HTML]{FDFFFD}} \color[HTML]{000000} 70.0 & {\cellcolor[HTML]{FDFFFD}} \color[HTML]{000000} 80.7 & {\cellcolor[HTML]{FDFFFD}} \color[HTML]{000000} 75.9 & {\cellcolor[HTML]{FDFFFD}} \color[HTML]{000000} 73.4 & {\cellcolor[HTML]{FDFFFD}} \color[HTML]{000000} 108. & {\cellcolor[HTML]{FDFFFD}} \color[HTML]{000000} 77.7 & {\cellcolor[HTML]{FDFFFD}} \color[HTML]{000000} 46.7 & {\cellcolor[HTML]{FDFFFD}} \color[HTML]{000000} 42.4 & {\cellcolor[HTML]{FDFFFD}} \color[HTML]{000000} 11.3 & {\cellcolor[HTML]{FDFFFD}} \color[HTML]{000000} 86.7 \\
\end{tabular}

\end{small}
}
\begin{small}

\caption{Reference-based and QE evaluation scores for greedy and MBR/QE decoding (1\textsuperscript{st} block), and ensembles (2\textsuperscript{nd} block), on en-so (FLORES200 test dataset). Higher scores are better, except MetricX, MetricX-QE, and TER, where lower is better. Green is better than greedy, red is worse.  Ensembles are defined in Table~\ref{tab:ensembles}. Significant differences from greedy (pairwise t-test) indicated by * for p<0.05, † for p<0.01, ‡ for p<0.001. The green diagonal in the 1\textsuperscript{st} block shows metrics prefer outputs from MBR/QE decoding using the same utility metric.}
\label{tab:subset_en_so_test}
\end{small}
    
\end{table}
\clearpage

\subsection{Results for English-Hindi (en-hi) on FLORES200 test dataset}
\begin{table}[h!]

\makebox[\textwidth]{
\setlength\tabcolsep{2 pt}
\begin{small}

\begin{tabular}{lccccccccccccccc}
\textbf{MBR/QE Method} \hfill \hfill \hfill \hfill \hfill \hfill \hfill \hfill \hfill \hfill \lrot{\textbf{Evaluated Metric}} & \lrot{MetricX} & \lrot{MetricX-QE} & \lrot{XCOMET-XXL} & \lrot{XCOMET-XL} & \lrot{CometKiwi23-XXL} & \lrot{CometKiwi23-XL} & \lrot{CometKiwi22} & \lrot{COMET22} & \lrot{IndicCOMET} & \lrot{BLEURT} & \lrot{YiSi} & \lrot{chrF} & \lrot{chrF++} & \lrot{sentBLEU} & \lrot{TER} \\
Greedy & {\cellcolor[HTML]{FDFFFD}} \color[HTML]{000000} 0.815 & {\cellcolor[HTML]{FDFFFD}} \color[HTML]{000000} 0.455 & {\cellcolor[HTML]{FDFFFD}} \color[HTML]{000000} 84.9 & {\cellcolor[HTML]{FDFFFD}} \color[HTML]{000000} 75.9 & {\cellcolor[HTML]{FDFFFD}} \color[HTML]{000000} 77.5 & {\cellcolor[HTML]{FDFFFD}} \color[HTML]{000000} 68.7 & {\cellcolor[HTML]{FDFFFD}} \color[HTML]{000000} 85.5 & {\cellcolor[HTML]{FDFFFD}} \color[HTML]{000000} 82.4 & {\cellcolor[HTML]{FDFFFD}} \color[HTML]{000000} 80.5 & {\cellcolor[HTML]{FDFFFD}} \color[HTML]{000000} 74.3 & {\cellcolor[HTML]{FDFFFD}} \color[HTML]{000000} 86.6 & {\cellcolor[HTML]{FDFFFD}} \color[HTML]{000000} 59.8 & {\cellcolor[HTML]{FDFFFD}} \color[HTML]{000000} 57.8 & {\cellcolor[HTML]{FDFFFD}} \color[HTML]{000000} 32.9 & {\cellcolor[HTML]{FDFFFD}} \color[HTML]{000000} 51.4 \\
\end{tabular}

\end{small}
}
\begin{small}

\caption{Reference-based and QE evaluation scores for greedy and MBR/QE decoding (1\textsuperscript{st} block), and ensembles (2\textsuperscript{nd} block), on en-hi (FLORES200 test dataset). Higher scores are better, except MetricX, MetricX-QE, and TER, where lower is better. Green is better than greedy, red is worse.  Ensembles are defined in Table~\ref{tab:ensembles}. Significant differences from greedy (pairwise t-test) indicated by * for p<0.05, † for p<0.01, ‡ for p<0.001. The green diagonal in the 1\textsuperscript{st} block shows metrics prefer outputs from MBR/QE decoding using the same utility metric.}
\label{tab:subset_en_hi_test}
\end{small}
    
\end{table}
\clearpage

\subsection{Results for English-Tamil (en-ta) on FLORES200 test dataset}
\begin{table}[h!]

\makebox[\textwidth]{
\setlength\tabcolsep{2 pt}
\begin{small}

\begin{tabular}{lccccccccccccccc}
\textbf{MBR/QE Method} \hfill \hfill \hfill \hfill \hfill \hfill \hfill \hfill \hfill \hfill \lrot{\textbf{Evaluated Metric}} & \lrot{MetricX} & \lrot{MetricX-QE} & \lrot{XCOMET-XXL} & \lrot{XCOMET-XL} & \lrot{CometKiwi23-XXL} & \lrot{CometKiwi23-XL} & \lrot{CometKiwi22} & \lrot{COMET22} & \lrot{IndicCOMET} & \lrot{BLEURT} & \lrot{YiSi} & \lrot{chrF} & \lrot{chrF++} & \lrot{sentBLEU} & \lrot{TER} \\
Greedy & {\cellcolor[HTML]{FDFFFD}} \color[HTML]{000000} 0.789 & {\cellcolor[HTML]{FDFFFD}} \color[HTML]{000000} 0.491 & {\cellcolor[HTML]{FDFFFD}} \color[HTML]{000000} 80.6 & {\cellcolor[HTML]{FDFFFD}} \color[HTML]{000000} 65.9 & {\cellcolor[HTML]{FDFFFD}} \color[HTML]{000000} 83.4 & {\cellcolor[HTML]{FDFFFD}} \color[HTML]{000000} 76.6 & {\cellcolor[HTML]{FDFFFD}} \color[HTML]{000000} 86.7 & {\cellcolor[HTML]{FDFFFD}} \color[HTML]{000000} 90.2 & {\cellcolor[HTML]{FDFFFD}} \color[HTML]{000000} 71.9 & {\cellcolor[HTML]{FDFFFD}} \color[HTML]{000000} 80.9 & {\cellcolor[HTML]{FDFFFD}} \color[HTML]{000000} 83.7 & {\cellcolor[HTML]{FDFFFD}} \color[HTML]{000000} 60.2 & {\cellcolor[HTML]{FDFFFD}} \color[HTML]{000000} 54.4 & {\cellcolor[HTML]{FDFFFD}} \color[HTML]{000000} 17.6 & {\cellcolor[HTML]{FDFFFD}} \color[HTML]{000000} 73.4 \\
\end{tabular}

\end{small}
}
\begin{small}

\caption{Reference-based and QE evaluation scores for greedy and MBR/QE decoding (1\textsuperscript{st} block), and ensembles (2\textsuperscript{nd} block), on en-ta (FLORES200 test dataset). Higher scores are better, except MetricX, MetricX-QE, and TER, where lower is better. Green is better than greedy, red is worse.  Ensembles are defined in Table~\ref{tab:ensembles}. Significant differences from greedy (pairwise t-test) indicated by * for p<0.05, † for p<0.01, ‡ for p<0.001. The green diagonal in the 1\textsuperscript{st} block shows metrics prefer outputs from MBR/QE decoding using the same utility metric.}
\label{tab:subset_en_ta_test}
\end{small}
    
\end{table}
\clearpage

\subsection{Results for English-Gujarati (en-gu) on FLORES200 test dataset}
\begin{table}[h!]

\makebox[\textwidth]{
\setlength\tabcolsep{2 pt}
\begin{small}

\begin{tabular}{lccccccccccccccc}
\textbf{MBR/QE Method} \hfill \hfill \hfill \hfill \hfill \hfill \hfill \hfill \hfill \hfill \lrot{\textbf{Evaluated Metric}} & \lrot{MetricX} & \lrot{MetricX-QE} & \lrot{XCOMET-XXL} & \lrot{XCOMET-XL} & \lrot{CometKiwi23-XXL} & \lrot{CometKiwi23-XL} & \lrot{CometKiwi22} & \lrot{COMET22} & \lrot{IndicCOMET} & \lrot{BLEURT} & \lrot{YiSi} & \lrot{chrF} & \lrot{chrF++} & \lrot{sentBLEU} & \lrot{TER} \\
Greedy & {\cellcolor[HTML]{FDFFFD}} \color[HTML]{000000} 0.794 & {\cellcolor[HTML]{FDFFFD}} \color[HTML]{000000} 0.263 & {\cellcolor[HTML]{FDFFFD}} \color[HTML]{000000} 94.3 & {\cellcolor[HTML]{FDFFFD}} \color[HTML]{000000} 86.2 & {\cellcolor[HTML]{FDFFFD}} \color[HTML]{000000} 76.2 & {\cellcolor[HTML]{FDFFFD}} \color[HTML]{000000} 69.1 & {\cellcolor[HTML]{FDFFFD}} \color[HTML]{000000} 87.7 & {\cellcolor[HTML]{FDFFFD}} \color[HTML]{000000} 89.1 & {\cellcolor[HTML]{FDFFFD}} \color[HTML]{000000} 95.2 & {\cellcolor[HTML]{FDFFFD}} \color[HTML]{000000} 83.2 & {\cellcolor[HTML]{FDFFFD}} \color[HTML]{000000} 84.7 & {\cellcolor[HTML]{FDFFFD}} \color[HTML]{000000} 54.9 & {\cellcolor[HTML]{FDFFFD}} \color[HTML]{000000} 51.7 & {\cellcolor[HTML]{FDFFFD}} \color[HTML]{000000} 21.8 & {\cellcolor[HTML]{FDFFFD}} \color[HTML]{000000} 65.0 \\
\end{tabular}

\end{small}
}
\begin{small}

\caption{Reference-based and QE evaluation scores for greedy and MBR/QE decoding (1\textsuperscript{st} block), and ensembles (2\textsuperscript{nd} block), on en-gu (FLORES200 test dataset). Higher scores are better, except MetricX, MetricX-QE, and TER, where lower is better. Green is better than greedy, red is worse.  Ensembles are defined in Table~\ref{tab:ensembles}. Significant differences from greedy (pairwise t-test) indicated by * for p<0.05, † for p<0.01, ‡ for p<0.001. The green diagonal in the 1\textsuperscript{st} block shows metrics prefer outputs from MBR/QE decoding using the same utility metric.}
\label{tab:subset_en_gu_test}
\end{small}
    
\end{table}
\clearpage

\subsection{Results for English-Malayalam (en-ml) on FLORES200 test dataset}
\begin{table}[h!]

\makebox[\textwidth]{
\setlength\tabcolsep{2 pt}
\begin{small}

\begin{tabular}{lccccccccccccccc}
\textbf{MBR/QE Method} \hfill \hfill \hfill \hfill \hfill \hfill \hfill \hfill \hfill \hfill \lrot{\textbf{Evaluated Metric}} & \lrot{MetricX} & \lrot{MetricX-QE} & \lrot{XCOMET-XXL} & \lrot{XCOMET-XL} & \lrot{CometKiwi23-XXL} & \lrot{CometKiwi23-XL} & \lrot{CometKiwi22} & \lrot{COMET22} & \lrot{IndicCOMET} & \lrot{BLEURT} & \lrot{YiSi} & \lrot{chrF} & \lrot{chrF++} & \lrot{sentBLEU} & \lrot{TER} \\
Greedy & {\cellcolor[HTML]{FDFFFD}} \color[HTML]{000000} 0.849 & {\cellcolor[HTML]{FDFFFD}} \color[HTML]{000000} 0.393 & {\cellcolor[HTML]{FDFFFD}} \color[HTML]{000000} 91.7 & {\cellcolor[HTML]{FDFFFD}} \color[HTML]{000000} 83.6 & {\cellcolor[HTML]{FDFFFD}} \color[HTML]{000000} 82.6 & {\cellcolor[HTML]{FDFFFD}} \color[HTML]{000000} 77.1 & {\cellcolor[HTML]{FDFFFD}} \color[HTML]{000000} 86.9 & {\cellcolor[HTML]{FDFFFD}} \color[HTML]{000000} 88.5 & {\cellcolor[HTML]{FDFFFD}} \color[HTML]{000000} 94.3 & {\cellcolor[HTML]{FDFFFD}} \color[HTML]{000000} 80.9 & {\cellcolor[HTML]{FDFFFD}} \color[HTML]{000000} 82.4 & {\cellcolor[HTML]{FDFFFD}} \color[HTML]{000000} 56.5 & {\cellcolor[HTML]{FDFFFD}} \color[HTML]{000000} 50.9 & {\cellcolor[HTML]{FDFFFD}} \color[HTML]{000000} 15.9 & {\cellcolor[HTML]{FDFFFD}} \color[HTML]{000000} 74.9 \\
\end{tabular}

\end{small}
}
\begin{small}

\caption{Reference-based and QE evaluation scores for greedy and MBR/QE decoding (1\textsuperscript{st} block), and ensembles (2\textsuperscript{nd} block), on en-ml (FLORES200 test dataset). Higher scores are better, except MetricX, MetricX-QE, and TER, where lower is better. Green is better than greedy, red is worse.  Ensembles are defined in Table~\ref{tab:ensembles}. Significant differences from greedy (pairwise t-test) indicated by * for p<0.05, † for p<0.01, ‡ for p<0.001. The green diagonal in the 1\textsuperscript{st} block shows metrics prefer outputs from MBR/QE decoding using the same utility metric.}
\label{tab:subset_en_ml_test}
\end{small}
    
\end{table}
\clearpage

\subsection{Results for English-Vietnamese (en-vi) on FLORES200 test dataset}
\begin{table}[h!]

\makebox[\textwidth]{
\setlength\tabcolsep{2 pt}
\begin{small}

\begin{tabular}{lcccccccccccccc}
\textbf{MBR/QE Method} \hfill \hfill \hfill \hfill \hfill \hfill \hfill \hfill \hfill \hfill \lrot{\textbf{Evaluated Metric}} & \lrot{MetricX} & \lrot{MetricX-QE} & \lrot{XCOMET-XXL} & \lrot{XCOMET-XL} & \lrot{CometKiwi23-XXL} & \lrot{CometKiwi23-XL} & \lrot{CometKiwi22} & \lrot{COMET22} & \lrot{BLEURT} & \lrot{YiSi} & \lrot{chrF} & \lrot{chrF++} & \lrot{sentBLEU} & \lrot{TER} \\
Greedy & {\cellcolor[HTML]{FDFFFD}} \color[HTML]{000000} 1.16 & {\cellcolor[HTML]{FDFFFD}} \color[HTML]{000000} 0.555 & {\cellcolor[HTML]{FDFFFD}} \color[HTML]{000000} 93.7 & {\cellcolor[HTML]{FDFFFD}} \color[HTML]{000000} 92.7 & {\cellcolor[HTML]{FDFFFD}} \color[HTML]{000000} 91.6 & {\cellcolor[HTML]{FDFFFD}} \color[HTML]{000000} 79.6 & {\cellcolor[HTML]{FDFFFD}} \color[HTML]{000000} 85.9 & {\cellcolor[HTML]{FDFFFD}} \color[HTML]{000000} 90.4 & {\cellcolor[HTML]{FDFFFD}} \color[HTML]{000000} 76.8 & {\cellcolor[HTML]{FDFFFD}} \color[HTML]{000000} 89.2 & {\cellcolor[HTML]{FDFFFD}} \color[HTML]{000000} 62.6 & {\cellcolor[HTML]{FDFFFD}} \color[HTML]{000000} 62.6 & {\cellcolor[HTML]{FDFFFD}} \color[HTML]{000000} 42.7 & {\cellcolor[HTML]{FDFFFD}} \color[HTML]{000000} 41.3 \\
\end{tabular}

\end{small}
}
\begin{small}

\caption{Reference-based and QE evaluation scores for greedy and MBR/QE decoding (1\textsuperscript{st} block), and ensembles (2\textsuperscript{nd} block), on en-vi (FLORES200 test dataset). Higher scores are better, except MetricX, MetricX-QE, and TER, where lower is better. Green is better than greedy, red is worse.  Ensembles are defined in Table~\ref{tab:ensembles}. Significant differences from greedy (pairwise t-test) indicated by * for p<0.05, † for p<0.01, ‡ for p<0.001. The green diagonal in the 1\textsuperscript{st} block shows metrics prefer outputs from MBR/QE decoding using the same utility metric.}
\label{tab:subset_en_vi_test}
\end{small}
    
\end{table}
\clearpage

\subsection{Results for English-Hungarian (en-hu) on FLORES200 test dataset}
\begin{table}[h!]

\makebox[\textwidth]{
\setlength\tabcolsep{2 pt}
\begin{small}

\begin{tabular}{lcccccccccccccc}
\textbf{MBR/QE Method} \hfill \hfill \hfill \hfill \hfill \hfill \hfill \hfill \hfill \hfill \lrot{\textbf{Evaluated Metric}} & \lrot{MetricX} & \lrot{MetricX-QE} & \lrot{XCOMET-XXL} & \lrot{XCOMET-XL} & \lrot{CometKiwi23-XXL} & \lrot{CometKiwi23-XL} & \lrot{CometKiwi22} & \lrot{COMET22} & \lrot{BLEURT} & \lrot{YiSi} & \lrot{chrF} & \lrot{chrF++} & \lrot{sentBLEU} & \lrot{TER} \\
Greedy & {\cellcolor[HTML]{FDFFFD}} \color[HTML]{000000} 0.589 & {\cellcolor[HTML]{FDFFFD}} \color[HTML]{000000} 0.338 & {\cellcolor[HTML]{FDFFFD}} \color[HTML]{000000} 96.2 & {\cellcolor[HTML]{FDFFFD}} \color[HTML]{000000} 94.1 & {\cellcolor[HTML]{FDFFFD}} \color[HTML]{000000} 93.7 & {\cellcolor[HTML]{FDFFFD}} \color[HTML]{000000} 82.7 & {\cellcolor[HTML]{FDFFFD}} \color[HTML]{000000} 87.9 & {\cellcolor[HTML]{FDFFFD}} \color[HTML]{000000} 90.9 & {\cellcolor[HTML]{FDFFFD}} \color[HTML]{000000} 90.5 & {\cellcolor[HTML]{FDFFFD}} \color[HTML]{000000} 84.6 & {\cellcolor[HTML]{FDFFFD}} \color[HTML]{000000} 60.6 & {\cellcolor[HTML]{FDFFFD}} \color[HTML]{000000} 57.4 & {\cellcolor[HTML]{FDFFFD}} \color[HTML]{000000} 26.7 & {\cellcolor[HTML]{FDFFFD}} \color[HTML]{000000} 58.7 \\
\end{tabular}

\end{small}
}
\begin{small}

\caption{Reference-based and QE evaluation scores for greedy and MBR/QE decoding (1\textsuperscript{st} block), and ensembles (2\textsuperscript{nd} block), on en-hu (FLORES200 test dataset). Higher scores are better, except MetricX, MetricX-QE, and TER, where lower is better. Green is better than greedy, red is worse.  Ensembles are defined in Table~\ref{tab:ensembles}. Significant differences from greedy (pairwise t-test) indicated by * for p<0.05, † for p<0.01, ‡ for p<0.001. The green diagonal in the 1\textsuperscript{st} block shows metrics prefer outputs from MBR/QE decoding using the same utility metric.}
\label{tab:subset_en_hu_test}
\end{small}
    
\end{table}
\clearpage

\subsection{Results for English-German (en-de) on WMT2023 dataset}
\begin{table}[h!]

\makebox[\textwidth]{
\setlength\tabcolsep{2 pt}
\begin{small}

\begin{tabular}{lcccccccccccccc}
\textbf{MBR/QE Method} \hfill \hfill \hfill \hfill \hfill \hfill \hfill \hfill \hfill \hfill \lrot{\textbf{Evaluated Metric}} & \lrot{MetricX} & \lrot{MetricX-QE} & \lrot{XCOMET-XXL} & \lrot{XCOMET-XL} & \lrot{CometKiwi23-XXL} & \lrot{CometKiwi23-XL} & \lrot{CometKiwi22} & \lrot{COMET22} & \lrot{BLEURT} & \lrot{YiSi} & \lrot{chrF} & \lrot{chrF++} & \lrot{sentBLEU} & \lrot{TER} \\
Greedy & {\cellcolor[HTML]{FDFFFD}} \color[HTML]{000000} 1.24 & {\cellcolor[HTML]{FDFFFD}} \color[HTML]{000000} 1.42 & {\cellcolor[HTML]{FDFFFD}} \color[HTML]{000000} 90.1 & {\cellcolor[HTML]{FDFFFD}} \color[HTML]{000000} 87.2 & {\cellcolor[HTML]{FDFFFD}} \color[HTML]{000000} 79.6 & {\cellcolor[HTML]{FDFFFD}} \color[HTML]{000000} 70.7 & {\cellcolor[HTML]{FDFFFD}} \color[HTML]{000000} 81.3 & {\cellcolor[HTML]{FDFFFD}} \color[HTML]{000000} 85.6 & {\cellcolor[HTML]{FDFFFD}} \color[HTML]{000000} 73.5 & {\cellcolor[HTML]{FDFFFD}} \color[HTML]{000000} 87.9 & {\cellcolor[HTML]{FDFFFD}} \color[HTML]{000000} 70.1 & {\cellcolor[HTML]{FDFFFD}} \color[HTML]{000000} 68.2 & {\cellcolor[HTML]{FDFFFD}} \color[HTML]{000000} 45.4 & {\cellcolor[HTML]{FDFFFD}} \color[HTML]{000000} 42.1 \\
\end{tabular}

\end{small}
}
\begin{small}

\caption{Reference-based and QE evaluation scores for greedy and MBR/QE decoding (1\textsuperscript{st} block), and ensembles (2\textsuperscript{nd} block), on en-de (WMT2023 dataset). Higher scores are better, except MetricX, MetricX-QE, and TER, where lower is better. Green is better than greedy, red is worse.  Ensembles are defined in Table~\ref{tab:ensembles}. Significant differences from greedy (pairwise t-test) indicated by * for p<0.05, † for p<0.01, ‡ for p<0.001. The green diagonal in the 1\textsuperscript{st} block shows metrics prefer outputs from MBR/QE decoding using the same utility metric.}
\label{tab:subset_en_de_test}
\end{small}
    
\end{table}
\clearpage

\subsection{Results for German-English (de-en) on WMT2023 dataset}
\begin{table}[h!]

\makebox[\textwidth]{
\setlength\tabcolsep{2 pt}
\begin{small}

\begin{tabular}{lcccccccccccccc}
\textbf{MBR/QE Method} \hfill \hfill \hfill \hfill \hfill \hfill \hfill \hfill \hfill \hfill \lrot{\textbf{Evaluated Metric}} & \lrot{MetricX} & \lrot{MetricX-QE} & \lrot{XCOMET-XXL} & \lrot{XCOMET-XL} & \lrot{CometKiwi23-XXL} & \lrot{CometKiwi23-XL} & \lrot{CometKiwi22} & \lrot{COMET22} & \lrot{BLEURT} & \lrot{YiSi} & \lrot{chrF} & \lrot{chrF++} & \lrot{sentBLEU} & \lrot{TER} \\
Greedy & {\cellcolor[HTML]{FDFFFD}} \color[HTML]{000000} 2.00 & {\cellcolor[HTML]{FDFFFD}} \color[HTML]{000000} 1.82 & {\cellcolor[HTML]{FDFFFD}} \color[HTML]{000000} 87.3 & {\cellcolor[HTML]{FDFFFD}} \color[HTML]{000000} 89.0 & {\cellcolor[HTML]{FDFFFD}} \color[HTML]{000000} 76.5 & {\cellcolor[HTML]{FDFFFD}} \color[HTML]{000000} 68.4 & {\cellcolor[HTML]{FDFFFD}} \color[HTML]{000000} 79.3 & {\cellcolor[HTML]{FDFFFD}} \color[HTML]{000000} 85.4 & {\cellcolor[HTML]{FDFFFD}} \color[HTML]{000000} 74.7 & {\cellcolor[HTML]{FDFFFD}} \color[HTML]{000000} 88.8 & {\cellcolor[HTML]{FDFFFD}} \color[HTML]{000000} 68.1 & {\cellcolor[HTML]{FDFFFD}} \color[HTML]{000000} 66.5 & {\cellcolor[HTML]{FDFFFD}} \color[HTML]{000000} 46.0 & {\cellcolor[HTML]{FDFFFD}} \color[HTML]{000000} 39.5 \\
\end{tabular}

\end{small}
}
\begin{small}

\caption{Reference-based and QE evaluation scores for greedy and MBR/QE decoding (1\textsuperscript{st} block), and ensembles (2\textsuperscript{nd} block), on de-en (WMT2023 dataset). Higher scores are better, except MetricX, MetricX-QE, and TER, where lower is better. Green is better than greedy, red is worse.  Ensembles are defined in Table~\ref{tab:ensembles}. Significant differences from greedy (pairwise t-test) indicated by * for p<0.05, † for p<0.01, ‡ for p<0.001. The green diagonal in the 1\textsuperscript{st} block shows metrics prefer outputs from MBR/QE decoding using the same utility metric.}
\label{tab:subset_de_en_test}
\end{small}
    
\end{table}
\clearpage

\subsection{Results for English-Chinese (en-zh) on WMT2023 dataset}
\begin{table}[h!]

\makebox[\textwidth]{
\setlength\tabcolsep{2 pt}
\begin{small}

\begin{tabular}{lcccccccccccccc}
\textbf{MBR/QE Method} \hfill \hfill \hfill \hfill \hfill \hfill \hfill \hfill \hfill \hfill \lrot{\textbf{Evaluated Metric}} & \lrot{MetricX} & \lrot{MetricX-QE} & \lrot{XCOMET-XXL} & \lrot{XCOMET-XL} & \lrot{CometKiwi23-XXL} & \lrot{CometKiwi23-XL} & \lrot{CometKiwi22} & \lrot{COMET22} & \lrot{BLEURT} & \lrot{YiSi} & \lrot{chrF} & \lrot{chrF++} & \lrot{sentBLEU} & \lrot{TER} \\
Greedy & {\cellcolor[HTML]{FDFFFD}} \color[HTML]{000000} 1.36 & {\cellcolor[HTML]{FDFFFD}} \color[HTML]{000000} 1.24 & {\cellcolor[HTML]{FDFFFD}} \color[HTML]{000000} 89.4 & {\cellcolor[HTML]{FDFFFD}} \color[HTML]{000000} 85.9 & {\cellcolor[HTML]{FDFFFD}} \color[HTML]{000000} 75.5 & {\cellcolor[HTML]{FDFFFD}} \color[HTML]{000000} 70.1 & {\cellcolor[HTML]{FDFFFD}} \color[HTML]{000000} 80.2 & {\cellcolor[HTML]{FDFFFD}} \color[HTML]{000000} 87.0 & {\cellcolor[HTML]{FDFFFD}} \color[HTML]{000000} 73.3 & {\cellcolor[HTML]{FDFFFD}} \color[HTML]{000000} 88.0 & {\cellcolor[HTML]{FDFFFD}} \color[HTML]{000000} 46.9 & {\cellcolor[HTML]{FDFFFD}} \color[HTML]{000000} 41.0 & {\cellcolor[HTML]{FDFFFD}} \color[HTML]{000000} 11.6 & {\cellcolor[HTML]{FDFFFD}} \color[HTML]{000000} 97.1 \\
\end{tabular}

\end{small}
}
\begin{small}

\caption{Reference-based and QE evaluation scores for greedy and MBR/QE decoding (1\textsuperscript{st} block), and ensembles (2\textsuperscript{nd} block), on en-zh (WMT2023 dataset). Higher scores are better, except MetricX, MetricX-QE, and TER, where lower is better. Green is better than greedy, red is worse.  Ensembles are defined in Table~\ref{tab:ensembles}. Significant differences from greedy (pairwise t-test) indicated by * for p<0.05, † for p<0.01, ‡ for p<0.001. The green diagonal in the 1\textsuperscript{st} block shows metrics prefer outputs from MBR/QE decoding using the same utility metric.}
\label{tab:subset_en_zh_test}
\end{small}
    
\end{table}
\clearpage

\subsection{Results for Chinese-English (zh-en) on WMT2023 dataset}
\begin{table}[h!]

\makebox[\textwidth]{
\setlength\tabcolsep{2 pt}
\begin{small}

\begin{tabular}{lcccccccccccccc}
\textbf{MBR/QE Method} \hfill \hfill \hfill \hfill \hfill \hfill \hfill \hfill \hfill \hfill \lrot{\textbf{Evaluated Metric}} & \lrot{MetricX} & \lrot{MetricX-QE} & \lrot{XCOMET-XXL} & \lrot{XCOMET-XL} & \lrot{CometKiwi23-XXL} & \lrot{CometKiwi23-XL} & \lrot{CometKiwi22} & \lrot{COMET22} & \lrot{BLEURT} & \lrot{YiSi} & \lrot{chrF} & \lrot{chrF++} & \lrot{sentBLEU} & \lrot{TER} \\
Greedy & {\cellcolor[HTML]{FDFFFD}} \color[HTML]{000000} 2.19 & {\cellcolor[HTML]{FDFFFD}} \color[HTML]{000000} 2.00 & {\cellcolor[HTML]{FDFFFD}} \color[HTML]{000000} 90.9 & {\cellcolor[HTML]{FDFFFD}} \color[HTML]{000000} 88.1 & {\cellcolor[HTML]{FDFFFD}} \color[HTML]{000000} 78.4 & {\cellcolor[HTML]{FDFFFD}} \color[HTML]{000000} 70.3 & {\cellcolor[HTML]{FDFFFD}} \color[HTML]{000000} 79.6 & {\cellcolor[HTML]{FDFFFD}} \color[HTML]{000000} 82.7 & {\cellcolor[HTML]{FDFFFD}} \color[HTML]{000000} 71.4 & {\cellcolor[HTML]{FDFFFD}} \color[HTML]{000000} 82.5 & {\cellcolor[HTML]{FDFFFD}} \color[HTML]{000000} 55.0 & {\cellcolor[HTML]{FDFFFD}} \color[HTML]{000000} 52.4 & {\cellcolor[HTML]{FDFFFD}} \color[HTML]{000000} 26.2 & {\cellcolor[HTML]{FDFFFD}} \color[HTML]{000000} 65.2 \\
\end{tabular}

\end{small}
}
\begin{small}

\caption{Reference-based and QE evaluation scores for greedy and MBR/QE decoding (1\textsuperscript{st} block), and ensembles (2\textsuperscript{nd} block), on zh-en (WMT2023 dataset). Higher scores are better, except MetricX, MetricX-QE, and TER, where lower is better. Green is better than greedy, red is worse.  Ensembles are defined in Table~\ref{tab:ensembles}. Significant differences from greedy (pairwise t-test) indicated by * for p<0.05, † for p<0.01, ‡ for p<0.001. The green diagonal in the 1\textsuperscript{st} block shows metrics prefer outputs from MBR/QE decoding using the same utility metric.}
\label{tab:subset_zh_en_test}
\end{small}
    
\end{table}
\clearpage


\end{document}